\documentclass[twoside]{article}

\usepackage[accepted]{aistats2023}
\usepackage[round]{natbib}

\usepackage[utf8]{inputenc} 
\usepackage[T1]{fontenc}    
\usepackage{hyperref}       
\usepackage{url}            
\usepackage{booktabs}       
\usepackage{amsfonts}       
\usepackage{nicefrac}       
\usepackage{microtype}      
\usepackage{xcolor}         
\usepackage{graphicx}       
\usepackage{float}          

\usepackage{amsmath,amssymb,amsthm}
\usepackage{centernot}
\newcommand{\R}{\mathbb{R}}

\newcommand{\E}{\mathbb{E}}
\newcommand{\Prob}{\mathbb{P}}
\newcommand{\para}[1]{{\it #1.} }
\newcommand{\rd}{\mathrm{d}}

\DeclareMathOperator*{\argmin}{\arg\!\min}
\DeclareMathOperator*{\rank}{rank}

\theoremstyle{definition}
\newtheorem{lemma}{Lemma}
\newtheorem{proposition}[lemma]{Proposition}

\newtheorem{theorem}[lemma]{Theorem}
\newtheorem{definition}[lemma]{Definition}

\begin{document}

%

%

\twocolumn[

\aistatstitle{Implications of sparsity and high triangle density for graph representation learning}

\aistatsauthor{ Hannah Sansford \And Alexander Modell \And  Nick Whiteley \And Patrick Rubin-Delanchy }

\aistatsaddress{ University of Bristol \And  University of Bristol \And University of Bristol \And University of Bristol } ]

\begin{abstract}
Recent work has shown that sparse graphs containing many triangles cannot be reproduced using a finite-dimensional representation of the nodes, in which link probabilities are inner products. Here, we show that such graphs can be reproduced using an infinite-dimensional inner product model, where the node representations lie on a low-dimensional manifold. Recovering a global representation of the manifold is impossible in a sparse regime. However, we can zoom in on local neighbourhoods, where a lower-dimensional representation is possible. As our constructions allow the points to be uniformly distributed on the manifold, we find evidence against the common perception that triangles imply community structure. 
\end{abstract}

\section{INTRODUCTION}
Network analysis is a common activity in many areas of science and technology, and has diverse goals as a result. Sometimes, it is of interest to estimate just a few, salient, parameters which generally drive connectivity, for example to inform policy (as the `R' number did, in Britain, during the COVID-19 pandemic \citep{rnumber}). A considerable amount of research has gone into developing simple models, such as the Erd\"os-Renyi graph, preferential attachment \citep{barabasi1999emergence}, or the Chung-Lu model \citep{chung2002average}, to explain global properties of real-world networks. However, in machine-learning, the goal is often to uncover hidden structure or make predictions about the nodes or edges, and these simple models are often felt to say too little about the data. Instead, the paradigm of graph embedding has emerged, in which each node is represented as a vector. There are several theoretical reasons to do this, but the most straightforward is convenience, including compatibility with existing machine-learning techniques, which often take feature vectors as input.

In the design of graph embedding algorithms, providing a realistic generative model for the data has not always been high priority. In fact, in some of the most popular approaches, such as DeepWalk \citep{perozzi2014deepwalk} or Node2vec \citep{grover2016node2vec}, there is no obvious generative model. But this is a problem for many applications, particularly in unsupervised settings, as there is no straightforward measure of goodness-of-fit. Moreover, data retention and privacy issues have increased the need for synthetic graph generation techniques. It would be useful to be able to harness the power of graph embeddings to reproduce rich structure present in real graphs, \emph{as well as their global properties}. 

There are approaches to graph embedding which do purport to model the data. Of particular interest here is the highly studied random dot product graph (RDPG) \citep{young2007random,athreya2017statistical}, in which the nodes each have a latent position $X_i \in \R^d$, which we aim to estimate via graph embedding, and the inner product $\langle X_i, X_j\rangle$ provides the corresponding connection probability. The model has been particularly useful for understanding graph spectral embedding, where the node representations are obtained from the eigenvectors of some matrix representation of graph, such as the normalised Laplacian \citep{tang2018limit}. However, a recent result has put its adequacy as a \emph{generative model} in doubt. The random dot product graph is unable to produce graphs in which the nodes make few connections but still form large numbers of triangles \citep{seshadhri2020impossibility}. These properties, known as sparsity and transitivity, are regarded as ubiquitous in network science \citep{newman2018networks} and are of course consistent with our own experience of social networks: most of us have few connections, relative to the size of the network and yet, often, `a friend of a friend is a friend'.

A follow-on paper \citep{chanpuriya2020node}, which proved an inspiration to us, was able to reproduce sparsity and high triangle density (formal definition to come), in theory, by modifying the connection probability to $\max\{0, \min(\langle X_i,Y_j\rangle, 1)\}$ and, in practice, using the kernel $\text{logistic}(\langle X_i,Y_j\rangle)$ (the latter already in common use), assigning two latent positions, $X_i$ and $Y_i$, to each node $i$. More recently, \cite{stolman2022classic} have attained the properties of interest using the softmax kernel over a low-dimensional embedding.


It is important to appreciate that, in our approach, the inner product is not an arbitrary kernel. It provides a default choice, because under mild assumptions it can be used to represent any other positive definite kernel, although typically in infinite dimension, by Mercer's theorem, known in machine-learning as the `kernel trick' \citep{steinwart2008support}. We find a transparent relationship between the eigenvalues of the kernel and the triangle density (see the supplementary material --- Lemma~\ref{lem:sum_lambda}), which shows negative eigenvalues can only \emph{reduce} the expected number of triangles in the graph. As a result, if a carefully crafted kernel seems to reproduce high triangle density \citep{chanpuriya2020node}, and negative eigenvalues aren't the reason, the likely problem with the RDPG is with its finite-dimension assumption. 


There would be little hope for practical inference in infinite dimension without structure. Instead, this paper shows that \emph{sparse graphs with high triangle density can be modelled using an infinite-dimensional inner product model, with representations on a low dimensional manifold}.


An implication of this manifold structure, that we later explore in depth, is the plausibility of global versus local embedding. Under levels of sparsity which are generally deemed reasonable, the manifold does not get populated everywhere, and perfect reconstruction is impossible. However, since a $d$-dimensional manifold (informally) locally resembles $d$-dimensional Euclidean space, we can `zoom in' to the subgraph containing nodes in a small neighbourhood of the latent space, and form an embedding which represents a flat approximation of the manifold over that region. This encompasses, but generalises, the notion of finding local community structure \citep{rohe2013blessing,spielman2013local,chen2020targeted}. 
A key technical contribution of our work is to construct manifolds for which we can simultaneously control: sparsity, triangle density, and the on-manifold distribution. One point of the latter is that we can place a uniform distribution on the manifold, and still achieve sparse graphs with high triangle density. Because there is no reasonable notion of community structure in this case, we find evidence against the common perception that triangles imply community structure. 

\paragraph{Related work} Several earlier works have proposed infinite dimensional versions of the RDPG and/or the hypothesis that latent positions live on a low-dimensional manifold \citep{tang2013universally,athreya2018estimation,manifold_structure,lei2021network,whiteley2021matrix}. The main goal of this paper, to generatively achieve sparsity and high triangle density, was not considered in these papers and indeed they contain several assumptions which make the goal impossible, including a finite rank or trace-class kernel. The paper \citet{whiteley2021matrix} provides a crucial element of our theory, allowing a manifold construction for which we can control the on-manifold distribution. We principally focus on triangle density, rather than (the formal definition of) transitivity, to align our work with \citet{seshadhri2020impossibility}; however, our graphs also achieve high transitivity. 


\section{ACHIEVING HIGH TRIANGLE DENSITY IN SPARSE GRAPHS} \label{triangle_density}
\paragraph{Setup} We consider a sequence of simple undirected graphs, $\+A^{(n)}$, on nodes indexed $1, \ldots, n$. We make no distinction between the graph and its adjacency matrix, $\+A^{(n)}$, which is binary, $n \times n$, and symmetric, with the standard convention that $\+A^{(n)}_{ij} = 1$ if nodes $i$ and $j$ have an edge, and zero otherwise.

We will assign to each node, $i$, an infinite-dimensional latent position, $X_i^{(n)} \in \R^{\mathbb{N}}$, from a probability distribution $F_n$ whose support $\*M_n$ (the precise definition of which comes later) only allows inner products that are valid probabilities, that is,
\begin{equation}\langle x, y \rangle := \sum_{k=1}^{\infty} x_k y_k \in [0,1] \quad \text{for all $x,y \in \*M_n$.}\label{eq:validity}\end{equation}
For convenience, the superscript $(n)$ will be suppressed from the latent positions, but there is no correspondence between the latent positions of graphs of different sizes, i.e. $X_i$ associated with $\+A^{(n)}$ is not the same as $X_i$ associated with $\+A^{(m)}$, for $n \neq m$. 

\begin{definition}[Infinite-dimensional random dot product graph]
  Let $X_1, \ldots, X_n \overset{i.i.d.}{\sim}F_n$. The matrix $\+A^{(n)}$ represents an infinite-dimensional random dot product graph (IRDPG) if 
  \[\+A^{(n)} \overset{ind}{\sim} \text{Bernoulli}(\langle X_i, X_j\rangle ),\]
  conditional on $X_1, \ldots, X_n$, for $1 \leq i \leq j \leq n$.
\end{definition}
This model has appeared, at different levels of generality, in many previous works \citep{young2007random,tang2013universally,athreya2017statistical,lei2021network}, and encodes several assumptions which we review at the end of this section. Of particular interest here is the trade-off between two key properties: the graph sparsity factor, $\rho_n$, and triangle density, $\Delta_n$, where  \citep{he2015graphon,seshadhri2020impossibility}
\[n \rho_n = n \iint \langle x,y \rangle \rd F_n(x) \rd F_n(y);\]
\[ n \Delta_n = {n \choose 3} \iiint \langle x,y \rangle \langle y,z \rangle \langle z,x \rangle \rd F_n(x) \rd F_n(y)\rd F_n(z),\]
are respectively the graph's expected degree, $\E(\sum_j \+A_{ij}^{(n)})$, and expected number of triangles, $\E(\sum_{i \neq j \neq k} \+A_{ij}^{(n)} \+A_{jk}^{(n)} \+A_{ki}^{(n)})$.

\begin{definition}[Spectral embedding into $\R^d$]
  Given $\+A^{(n)}$ has eigendecomposition $\+A^{(n)} = \sum_i \lambda_i u_i u_i^\top$, its \emph{adjacency spectral embedding} into $d$ dimensions is given by the rows of
\begin{equation*}
    \hat{\+X} := (|\lambda_1|^{1/2} u_1, \ldots, |\lambda_d|^{1/2} u_d).
\end{equation*}
The point cloud $\hat{\+X}$ can be interpreted as a geometric representation of the best rank-$d$ approximation of $\+A^{(n)}$, in the sense that, provided $\lambda_1,\ldots,\lambda_d > 0$ (as is generally assumed throughout this paper),
\begin{equation*}
    \hat{\+X} = \argmin_{\+X : \rank(\+X) \leq d} \|\+X\+X^\top - \+A^{(n)}\|_\text{F}.
\end{equation*}
The matrix input may also be rectangular (e.g. the `core-periphery slice' (Section \ref{sec:implication} - local embedding)), in which case, singular values and left singular vectors replace eigenvalues and eigenvectors respectively.
\end{definition}

For the reader's convenience, we recall the standard rate-of-growth notation
\begin{align*}
  o(n)&: \text{`strictly slower than $n$';}\\
  O(n)&: \text{`slower than $n$';}\\
  \omega(n)&: \text{`strictly faster than $n$';}\\
  \Omega(n)&: \text{`faster than $n$';}\\
  \Theta(n)&: \text{`as fast as $n$', i.e., $O(n)$ and $\Omega(n)$.}
\end{align*}

\paragraph{Question}
Motivated by recent impossibility results concerning triangle densities under sparse finite-dimensional dot product models \citep{seshadhri2020impossibility}, the question which is the starting point for our contribution is: can a sparse IRDPG, i.e. $\rho_n = o(1)$, have high triangle density, i.e., $\Delta_n = \Omega(1)$?

\paragraph{Main result} The answer is yes. Moreover, in contrast to the results of \cite{seshadhri2020impossibility} which show it is impossible to have sparsity and high triangle density when  $F_n$ is supported on a finite-dimensional vector space, we show that sparsity and high triangle density \emph{is} possible when $F_n$ is supported on a low-dimensional manifold $\mathcal{M}_n$ embedded in an infinite-dimensional space. We can even obtain the extreme regime $\rho_n = \Theta(1/n), \Delta_n = \Theta(1)$, which includes asymptotically constant expected degree and triangle density.



\paragraph{Approach}
Our proof is by example: we construct $\mathcal{M}_n$ through a non-distortive homeomorphism $\phi_n$ applied to a compact topological manifold $\mathcal{Z}_n \subset \R^d$. In our examples, $\mathcal{Z}_n$ is either a growing interval or a growing circle. By non-distortive we mean: the Euclidean length of a curve $\eta(t)$ on $\*Z_n$ is equal to the (infinite dimensional) Euclidean length of the image of this curve $\phi_n\{\eta(t)\}$ on $\*M_n$. If $\*Z_n = [0,1]^2$, for example, we can think of a sheet of paper which is being bent or folded, but not cut or stretched. This property allows us to easily control the on-manifold distribution, an implication of which is given in Section \ref{sec:implication} (triangles and community structure). Moreover, the map $\phi_n$ is chosen in such a way that we can calculate the sparsity and triangle density of the resulting graph. The technical details of the construction are delegated to the supplementary material. 


\begin{figure*}[t]
\centering
\includegraphics[scale=0.4]{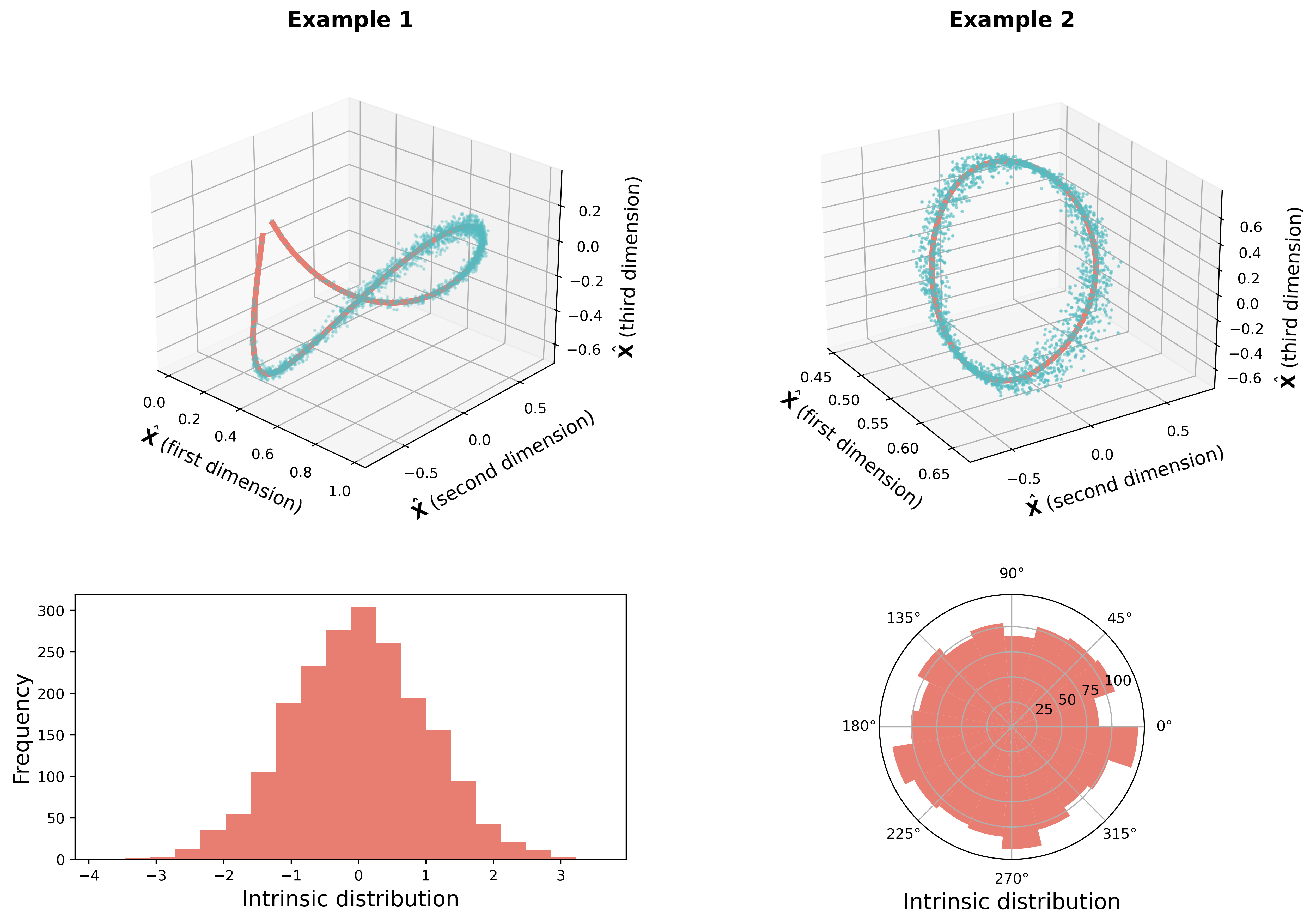}
\caption{The upper plots show the manifolds of our concrete examples (red curve) with the spectral embeddings (blue points) of a graph simulated from the model, and aligned using Procrustes analysis. The lower plots are histograms showing the intrinsic distribution of $X_i$, with $n=2000$, $\sigma_n=n/2000$ and $r_n = n/2000$.}
\label{fig:manifolds}
\end{figure*}

\paragraph{Example 1} Let $G_n$ be a Gaussian distribution with mean zero and variance $\sigma^2_n$, truncated to the interval $\*Z_n = [-t_n, t_n]$, for $t_n > 0$. We let $t_n = \sigma^2_n$ so that mass of $\mathcal{N}(0,\sigma_n^2)$ on the complement of $\mathcal{Z}_n$ tends to zero as $n\to\infty$. For this example:

\begin{lemma}
\label{GRBF_lemma}
There exists a non-distortive homeomorphism $\phi^{\text{RBF}}_n: \mathcal{Z}_n \to \R^\mathbb{N}$ such that for $\*M_n = \phi^{\text{RBF}}_n(\*Z_n)$ and with $F_n$ the pushforward by $\phi^{\text{RBF}}_n$ of the truncated Gaussian distribution $G_n$, 
 \begin{align*}
    \rho_n = \Theta(\sigma^{-1}_n), \quad \Delta_n = \Theta(n^2 \sigma^{-2}_n),
 \end{align*}
 so that $\rho_n = \Theta(1/n)$ and $\Delta_n = \Theta(1)$ if $\sigma_n = \Theta(n)$. 
\end{lemma}



\paragraph{Example 2} Let $G_n$ be the uniform distribution on the circle $\*Z_n = \{ x \in \mathbb{R}^2 : \|x\|=r_n \}$. For this example:

\begin{lemma}
  \label{sphere_lemma}
  There exists a non-distortive homeomorphism $\phi^\text{S}_n: \mathcal{Z}_n \to \R^\mathbb{N}$ such that for $\*M_n = \phi^\text{S}_n(\*Z_n)$ and $F_n$ the pushforward of $G_n$ by $\phi^\text{S}_n$, 
  \begin{align*}
    \rho_n = O(r^{-1}_n), \quad \Delta_n = \Omega(n^2 r^{-2}_n),
  \end{align*}
    so that $\rho_n = O(1/n)$ and $\Delta_n = \Omega(1)$ if $r_n = \Theta(n)$.
\end{lemma}


We note that in the context of Lemma~\ref{sphere_lemma}, $F_n$ is the uniform probability measure on $\*M_n$, since $G_n$ is uniform and $\phi^\text{S}_n$ is non-distortive. This is also an example where the expected degrees are identical and therefore bounded for each node, as in \citet{seshadhri2020impossibility}.

The two examples are illustrated in Figure~\ref{fig:manifolds}, with $n=2000$, $\sigma_n = n/2000$ and $r_n = n/2000$. \cite{seshadhri2020impossibility} note that for various real-world graphs,
the subgraphs containing nodes of degree $c \in [10,50]$ or less have triangle density greater than one (and in many cases greater than ten).
Our examples, and several others, are able to produce graphs with (constant) expected degrees and triangle densities in the same range, which we demonstrate in the supplementary materials. We also illustrate that the logistic model studied in  \cite{chanpuriya2020node}, and employed in Spotify's recommendation system \citep{johnson2014logistic}, forms a manifold in (indefinite) inner product space.

\paragraph{Model assumptions} Allowing self-loops is not crucial, and is simply done to make the analysis more reader-friendly. Forbidding self-loops will not change the expected number of triangles, and can only reduce the expected degree by at most one.

That the latent positions are i.i.d., edges conditionally independent, with probability given by an inner product, are testable assumptions. Given an embedding, $\hat X_1, \ldots, \hat X_n$, the first can be investigated using point process analysis techniques to detect dependence, such as repulsion \citep{baddeley2015spatial}. The second could be addressed using held out information, such as time, to uncover edge correlations not captured by $\hat X_1, \ldots, \hat X_n$. A recent paper has observed that edge independence can be problematic for reproducing triangle density while controlling a notion of overlap, for a fixed edge probability matrix \citep{chanpuriya2021power}. However, in our model, the edge probability matrix is random. Regarding the assumption of an inner product, the model cannot explain any large-magnitude negative eigenvalue in $\+A^{(n)}$.

These assumptions have been relaxed in many works  \citep{caron2017sparse,xie2021efficient,padilla2019change,GRDPG,lei2021network}. However, it is important to appreciate that they are appropriate for our goal: to show that sparsity and high triangle density are possible with node representations on a low-dimensional manifold, under a setup as close as possible to \cite{seshadhri2020impossibility}, contrasting their result. Moreover, if these properties are possible within a set of strong assumptions, they are possible more generally.

\paragraph{Relation to graphon models} Suppose the node representations are i.i.d. uniform on the unit interval, with connection probabilities given by an \textit{unknown} function $g: [0,1] \times [0,1] \to [0,1]$. If the graphon $g$ is fixed, sparse graphs are impossible since expected degree grows as $\Theta(n)$. If we scale $g$ by a sparsity factor, as is common in the literature \citep{lunde2019subsampling,lei2021network,wolfe2013graphon}, then high triangle density becomes impossible. In particular, if $n \rho_n = o(n^{1/3})$, then $\Delta_n = o(1)$. The proof is a straightforward application of the definition of $\Delta_n$ and can be found in the supplementary material. \cite{caron2017sparsity} prove that an extension of the graphon model, called \textit{graphex processes}, can simultaneously achieve sparsity and constant clustering coefficients.

\section{IMPLICATIONS}\label{sec:implication}

We now turn our attention away from the generative properties of the IRDPG, i.e. producing sparsity and high triangle density, to the implications of our model for performing inference on network data.

\paragraph{Global manifold learning} Most real world graphs are perceived to be very sparse, in that their average degree is much smaller than $n$, and it is common in network theory to equate this with the expected degree being $o(\log n)$ \citep{krivelevich2003largest, amini2013pseudo, le2017concentration}. The threshold $\log(n)$ has special significance because it is the rate under which an Erd\"os-R\'enyi graph is disconnected with high probability \citep{bollobas1998graduate}, and exact recovery of network communities is impossible \citep{abbe2017community}. The same threshold appears here, but for a different reason.

We can only expect to recover the manifold from $X_1, \ldots, X_n$ if they don't leave large parts of $\*M_n$ uncovered. We measure the largest gap using the Hausdorff distance
\[d_H(\{X_i\}, \*M_n) := \sup_{x \in \*M_n} \left\{\min_{i \in \{1, \ldots, n\}}d_{\circ}(X_i,x)\right\},\]
where $d_{\circ}(x,y)$ denotes geodesic distance on $\*M_n$, the length of the shortest curve between $x$ and $y$ on $\*M_n$.
\begin{lemma} \label{lem:gaps}
  If $\*M_n$ is homeomorphic to a closed interval, $F_n$ is uniform on $\*M_n$, and there exists $c>0$ such that $\|x\| \geq c$, for all $x \in \*M_n$, then
  \[d_H(\{X_i\}, \*M_n) \overset{p}{\rightarrow} 0 \Rightarrow n \rho_n = \Omega\{\log(n)\}.\]
\end{lemma} 
Given just $X_1, \ldots, X_n$, finding a $d$-dimensional representation of the nodes in which geodesic distances are faithfully reproduced is therefore impossible in general, unless $n \rho_n = \Omega\{\log(n)\}$. In practice, we only have access to estimates $\hat{X}_1, \ldots, \hat{X}_n$, which adds further complications to manifold learning. 

In fact, in this semi-sparse regime, there are problems relating to estimating $X_i$. For example, existing works on spectral embedding in the infinite-dimensional case make heavy use of the trace-class assumption \citep{tang2013universally,manifold_structure,lei2021network}, $\rho^{-1}_n\sum_{i=1}^\infty \lambda_i < \infty$, whereas $n \rho_n = O\{\log(n)\}$ and $\Delta_n = \Omega(1)$ require this sum to be unbounded in $n$.

\paragraph{Local embedding} When $\*M_n$ is a manifold, as it is in the examples above, we now explore the idea that we can `zoom in' to a particular location in latent space and faithfully represent the rich local structure with low-dimensional embeddings.

Given a ball with centre $x \in \*M_n \setminus \{0\}$ and radius $r$, denoted $B_{r}(x)$, we consider two local views of the graph: the \emph{core}, $\+A^{\text{C}}$, is the subgraph on the nodes with latent position in $B_{r}(x)$; the \emph{core-periphery}, $\+A^{\text{C-P}}$, is a graph on the full nodeset with adjacency matrix
\[\+A^{\text{C-P}}_{ij} = \begin{cases}\+A_{ij} &\text{if $X_i \in B_{r}(x)$ or $X_j \in B_{r}(x)$,}\\
      0 & \text{otherwise.}\end{cases}\]
In what follows, we will refer to a `slice' of this matrix, $\+A^{\text{C-P slice}} = \+A^{\text{C-P}}_{i: X_i \in B_r(x), j \in \{1, \ldots, n\}}$. 

We propose to find a low-rank approximate factorisation of these `local adjacency matrices', $\+A^{\text{C}}$ and $\+A^{\text{C-P slice}} $, and argue that, under certain conditions, this low-rank assumption is reasonable. We provide some theoretical reasons to prefer the core-periphery; however, our experiments with real data find them comparable.

To describe these graphs as $n$ grows, we have to assume the manifolds are nested, that is, $\*M_n \subseteq \*M_{n+1}$, for all $n$. We impose two other regularity conditions. First, $n / v_n \rightarrow \infty$, where $v_n = \int_{\mathcal{M}_n} 1 \rd x$ is the intrinsic volume of the manifold, which would preclude a constant expected degree in the examples of the previous section, but allows any faster rate. Second, there exists $R \in (0, \|x\|/\sqrt{6})$ and a constant $C>0$ such that \[\Prob\{X_i \in B_r(x)\} \geq \frac{C}{v_n} r^d, \quad \text{for all $r \leq R$}.\]

  
\begin{figure*}[t]
\centering
\includegraphics[width=0.8\linewidth]{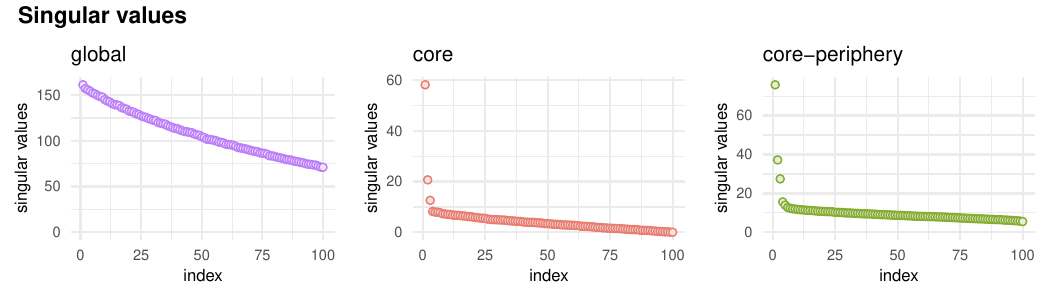}
\includegraphics[width=0.8\linewidth]{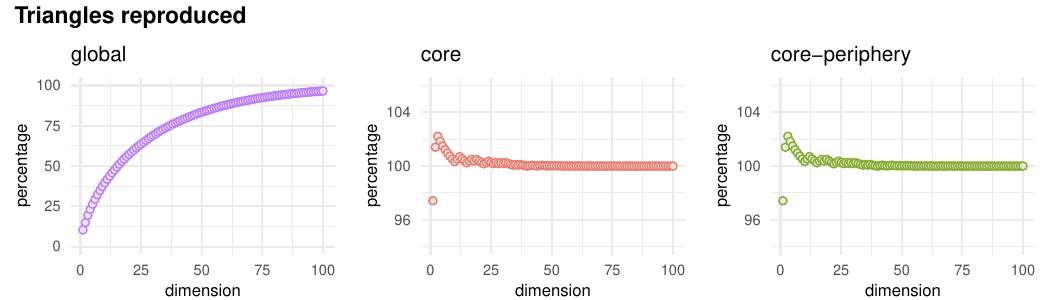}
\caption{Global versus local embedding. For a graph on 20,000 nodes on a 2-dimensional manifold, we estimate a 200-node neighbourhood around an arbitrary query node (details in main text). Top row: top 100 singular values of the graph, core, and core-periphery slice. The latter, but not the first, roughly distinguish three ($d+1$) large singular values, indicating the manifold dimension. Bottom row: percentage of triangles reproduced in simulations in the relevant graph using spectral embedding as the dimension increases. The triangle density of the core (middle and right) is well-explained using only a few dimensions (e.g. 1).}
\label{fig:svs_and_triangles}
\end{figure*}

  Under these assumptions, the core is \emph{dense}, with high triangle density.
  \begin{theorem}\label{thm:subgraph_density}
    Fix $x \in \*M_n \setminus \{0\}$ and let $r_n = \omega\{(v_n/n)^{1/d}\},  r_n \leq  R < \| x \|/\sqrt{6}$, which could be fixed or shrinking. The graph core corresponding to the ball $B_{r_n}(x)$ has $\Omega(n r^d/v_n)$ nodes in expectation, expected degree $\Omega(n r^d/v_n)$ and triangle density $\Omega(n r^d/v_n )^2$. 
  \end{theorem}
  If the manifold is smooth, a local neighbourhood $\*M_n \cap B_r(x)$ can be approximated up to arbitrary precision by a $d$-dimensional plane, the approximation improving as $r \rightarrow 0$. However, we can go too far, and collect very few points $X_i \in B_r(x)$. Theorem~\ref{thm:subgraph_density} tells us how $r$ should shrink with $n$ to avoid this.
  
  Under a flat approximation, \emph{the core and core-periphery slice have low rank}. To be precise, the matrices $\+P^{\text{C}} := \+P_{i,j \in \*C}$ and  $\+P^{\text{C-P slice}} := \+P_{i \in \*C, j \in \{1, \ldots, n\}}$, where $\+P = \langle X_i, X_j \rangle_{i, j \in \{1, \ldots, n\}}$ and $\*C = \{i: X_i \in B_{r}(x)\}$, have rank at most $d+1$. As a result, we can hope to recover local, finite-dimensional estimates of $X_i \in B_r(x)$ using approximate matrix factorisation of $\+A^{\text{C}}$ or $\+A^{\text{C-P slice}}$. Moreover, the approximate ranks of those matrices provide a local estimate of the manifold dimension.
  
  We find however that, given `noise-free' factorisations $\+P^{\text{C}} = \+X^{\text{C}} {(\+X^{\text{C}})}^\top$ and $\+P^{\text{C-P slice}} = \+X^{\text{C-P}} {(\+Y^{\text{C-P}})}^\top$, where $\+X^{\text{C}}, \+X^{\text{C-P}} \in \R^{|\*C| \times (d+1)}, \+Y^{\text{C-P}} \in \R^{n \times (d+1)}$, the original $X_i \in B_r(x)$ are linear images of the rows of $\+X^{\text{C-P}}$, but are not necessarily recoverable from $\+X^{\text{C}}$. 

 We have not discussed how a neighbourhood might be found in practice. There are many algorithms suited to this purpose \citep{haveliwala2003topic,spielman2004nearly,andersen2006local}, and even a simple similarity measure such as ``number of shared neighbours'' could be expected to be consistent \citep{chen2016statistical}. The supplementary material contains an experiment demonstrating that this measure forms a reasonable surrogate for the theoretical neighbourhood. There are two other ways a proxy for neighbourhood could be available. First, the graph may have associated covariates, such as time or location, which allow us to zoom into particular subgraphs (see classification experiment). Second, in many applications, a core-periphery graph is precisely all we have. For example, there is a real sense in which the famous Enron email dataset \citep{cohen2009enron} is a core-periphery view of the network of worldwide emails, focussed on a group of company executives at a period in time.
 
 \paragraph{Triangles and community structure}
It is known that certain forms of community structure, particularly when heterophilic (e.g. `opposites attract'), do not give rise to triangles \citep{newman2018networks}. For example, a bipartite graph has no triangles, but may still have well-defined communities. In other words,
  \[\text{community structure} \centernot \implies \text{high triangle density}.\]
  However, there is a perception that the reverse implication is true (see e.g.  \cite{durak2012degree, seshadhri2020impossibility} and references therein). 
   For example, the number of triangles divided by the number of connected triples is commonly known as the `clustering coefficient' and a high value of this statistic is considered a manifestation of community structure.
   
       To the contrary, our constructions achieve high triangle density, and high clustering coefficients (e.g. \textasciitilde 0.53 in Figure~\ref{fig:manifolds}, Example~2), when no sensible notion of community structure is present, for example with positions which are uniformly distributed on a manifold. In other words,
  \[ \text{high triangle density} \centernot \implies \text{community structure}.\]

 \section{EXPERIMENTS}
 
 \begin{figure*}[t]
\centering
\includegraphics[width=0.9\linewidth]{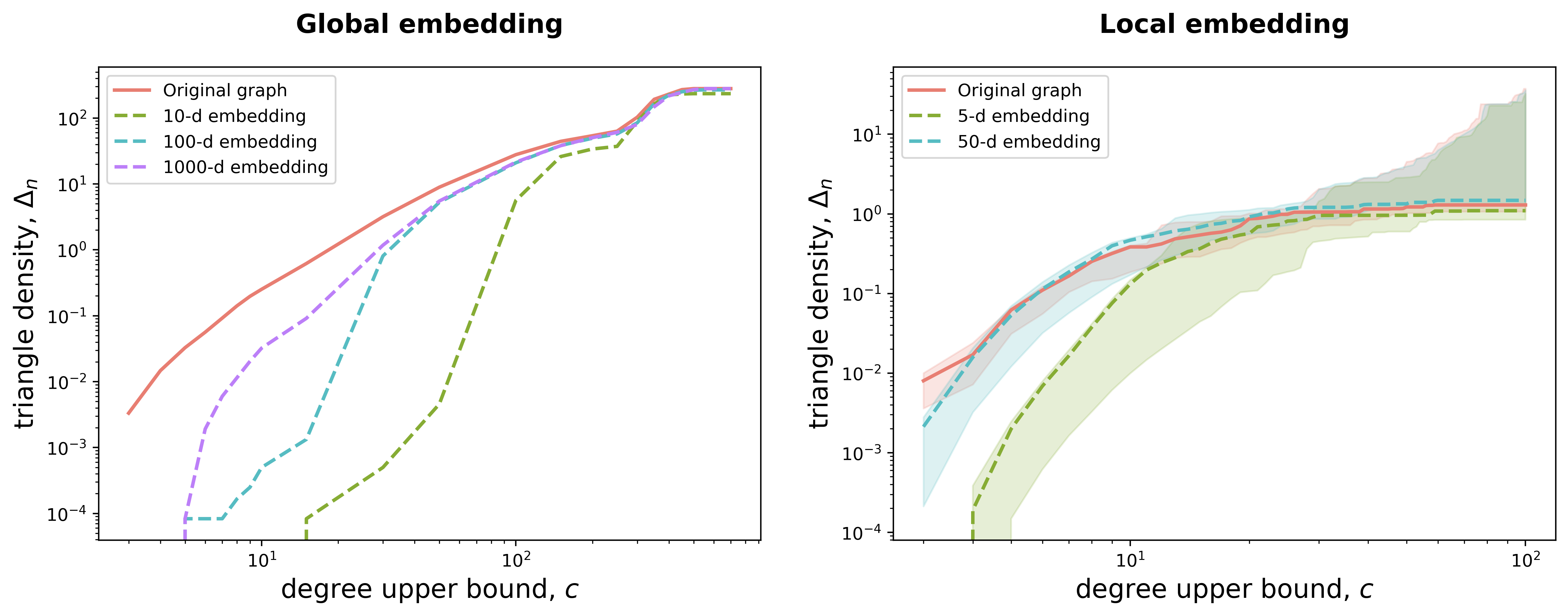}
\caption{True and recovered triangle counts on subgraphs of a High Energy Physics coauthorship network containing the nodes with degree, at most, $c$. The number of triangles is divided by size of the graph, $n$, so that it corresponds to $\Delta_n$. The local neighbourhood used to produce the right-hand plot is constructed by taking the 500 nodes with most links in common with an arbitrary `query' node. See the main text for details of how the curves and shaded regions are produced. Observe how the local embedding is able to recover triangle counts on much lower degree subgraphs than the global embedding, and with much lower dimensional embeddings.}
\label{fig:low_degree_triangles}
\end{figure*}
 
 \paragraph{Simulated graph} To compare local versus global embedding, we simulate a graph on a 2-dimensional manifold based on the $\phi_n^\text{RBF}$ map from Lemma \ref{GRBF_lemma}, setting $d = 2$, $n = 20,000$, and $G_n = \text{uniform}[-10,10]^2$. Picking an arbitrary `query' node, $i$, we construct a neighbourhood of the 200 nodes with the most links in common with $i$. We see a sharp `elbow' in the spectrum of the core and core-periphery graphs at $3 = d+1$, whereas there is no obvious cut-off point in the spectrum of the full graph (at least within the first 100 singular values, which are already quite expensive to compute). The triangle density of the core is well-approximated using core or core-periphery embeddings with only a few dimensions (even just 1), but the full graph requires a much higher dimension.
 
 To be clear, we do not claim that a global embedding is inferior to a local embedding, just that large geodesic distances in the former, irrelevant in the latter, must somehow be `made up' or grossly underestimated by Euclidean distance. On inspection of the central limits for spectral embedding \citep[Th. 4 and 7]{GRDPG}--- which we only treat as indicative because they assume a finite rank --- the error distributions of the core positions are the same, up to linear transformation, whether we embed the full graph or the core-periphery. By this (limited) evidence, there is no statistical gain (or loss) in embedding the core-periphery over the whole graph, when we are only interested in the core. The clear, practical advantage of the former is an enormous reduction in dimension.

\paragraph{Recovery of low-degree triangles} \cite{seshadhri2020impossibility} empirically validate their claims by demonstrating the inability of various (global) embeddings to recover the low-degree triangles found in real-world graphs. In Figure \ref{fig:low_degree_triangles}, we illustrate the ability of \textit{local} embeddings to capture this low-degree structure. Specifically, we demonstrate this on the dataset used in \cite{seshadhri2020impossibility}: a High Energy Physics coauthorship network with 12,008 nodes (ca-HepPh), obtained from the SNAP graph repository \citep{snapnets}. The plots show the triangle density ($y$-axis) among nodes with degree, at most, $c$ ($x$-axis). As in the simulated example, the local neighbourhood is constructed by sampling an arbitrary `query' node, $i$, and finding the 500 nodes that have most links in common with $i$. 

The dashed curves on the left-hand (global embedding) plot are the average triangle densities of 100 samples from embeddings of each dimension, while the solid red curve is the true triangle density of the original graph. On the right-hand (local embedding) plot, the solid and dashed curves are the 50th percentiles from 50 sampled neighbourhoods, and the shaded regions encompass the area between the 25th and 75th percentiles. For each sampled neighbourhood, the average triangle density is calculated from 100 graphs sampled from the corresponding embedding.

For an embedding of the core graph into only 5 dimensions, we accurately recover triangle counts on subgraphs with maximum degree greater than 10. Moreover, embedding the core graph into 50 dimensions recovers near-perfect triangle counts for all subgraphs. This is in contrast to the global embedding, which struggles to recover triangles on nodes with degree less than 100, even with a 1000-dimensional embedding. We expect that a graph containing fewer nodes will require a lower dimensional embedding to recover its structure, and hence triangles. However, looking at embedding dimension as a proportion of graph size, a 5-dimensional embedding on our local neighbourhood might be comparable to a 100-dimensional embedding on the entire graph. Likewise, a 50-dimensional embedding on the local graph might be comparable to a 1000-dimensional embedding on the full graph, which is evidently not the case.

\begin{figure*}[h!]
\centering
\includegraphics[width=0.9\linewidth]{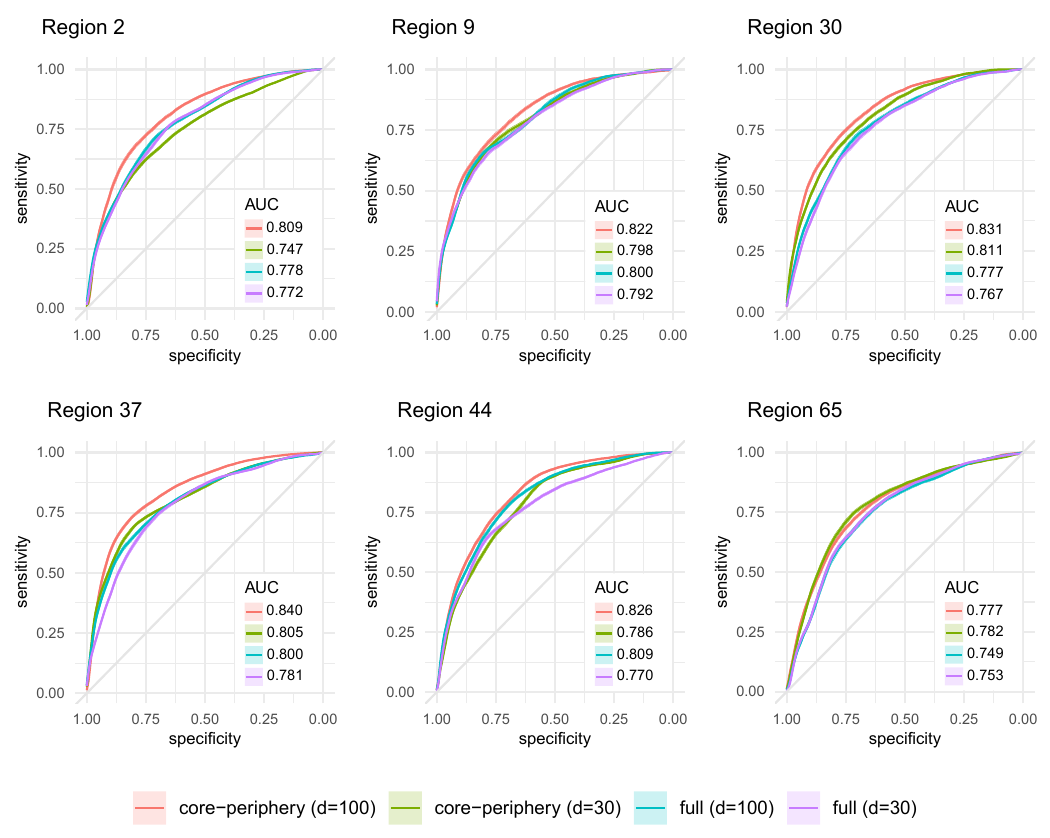}
\caption{Classification of tissue types from a diffusion MRI scan. For each region, we train a classifier to predict the tissue type of each node using the core-periphery and global embeddings into dimensions $d=30$ and $d=100$. For 100 Monte-Carlo iterations, a random sample of 20\% of the positions are held out for validation and the corresponding mean receiver-operating curves and mean area-under-the-curve (AUC) statistics are shown. Error bands displaying $\pm 2$ standard errors are shown, but are too small to be visible. The core-periphery embedding performs comparably to the global embedding, even with a lower dimension.
}
\label{fig:cp_rocs}
\end{figure*}

\paragraph{Classification}To demonstrate the merits of a local core-periphery embedding, we consider a dataset of a diffusion MRI scan of a connectome, first analysed in \cite{priebe2019two}. A graph is generated from the scan using the NeuroData MR Graph pipeline \citep{kiar2018high}, with vertices representing subregions defined via `spatial proximity and edges defined by tensor-based fiber streamlines connecting these regions' \citep{priebe2019two}, and we consider only the largest connected component, containing approximately 40,000 nodes. Each node is labelled as belonging to one of 70 regions of the brain and is also labelled with its tissue type: grey or white. 

For the six largest regions (each with between 1,351 and 2,316 nodes), we consider the Laplacian spectral embedding (see the supplementary materials for details) of the relevant nodes obtained using the full graph, and the corresponding core-periphery `slice' into both $d=30$ and $d=100$ dimensions. 100 and 30 were chosen as roughly logarithmic in the sizes of the full graph and local subgraphs, respectively, the rate required for Wasserstein consistency \citep{lei2021network}. To give an indication of compute time (which is a loose upper bound on all other experiments in this paper), the 100-dimensional embeddings for the core-periphery and full graphs take approximately 25 seconds and 397 seconds respectively on a 2017 quad-core MacBook Pro.  

We use the embeddings as the input to an $\ell_2$-regularised logistic regression classifier (the regularisation parameter for which is chosen via 10-fold cross-validation) to predict the tissue type label of each node. For 100 Monte-Carlo iterations, a random subset of 20\% of the embedding positions are held out for validation, and the rest are used to train the classifier. The mean receiver-operating curves on the validation sets are shown in Figure~\ref{fig:cp_rocs}, with bands showing $\pm 2$ standard errors.

It can be seen that, in general, the core-periphery embedding obtains a classification performance which is comparable to that of the global embedding, even with a lower-dimensional embedding. A comparison of core versus core-periphery classification performance is in the supplementary materials. 


\section{DISCUSSION}
This paper shows that sparse graphs with high triangle density are possible under an inner product model in which the node representations lie on a finite-dimensional manifold in infinite-dimensional space. We believe this is a simpler and more general answer to the corresponding impossibility result for inner product models in finite-dimensional space, compared to existing proposals such as the approach of \cite{chanpuriya2020node}, where each node has \emph{two representations}, $X_i$ and $Y_i$, and nodes $i$ and $j$ connect with probability $\text{logistic}(\langle X_i,Y_j\rangle)$' . A simpler model is easier to test, communicate, and reason about, and these are important advantages in network modelling applications affecting society, such as epidemiology. 


The presence of a manifold embedded in infinite-dimensional space would indicate several possibilities which we haven't yet pursued. For node classification, it is not actually obvious that graph embedding is necessary at all, just as dimension reduction does not always help regression \citep{jolliffe1982note}. In standard regression, there is an interesting line of research showing that several techniques, including feedforward neural networks \citep{nakada2020adaptive} and $K$ nearest neighbour methods \citep{kpotufe2011k}, automatically adapt to manifold structure in feature space. In a similar vein, classification algorithms working directly on the graph, i.e., without embedding, might be developed to exploit the intrinsic dimension of the nodes, on more informed assumptions about the manifold than were available previously. Another interesting direction for future work would be to investigate how one might combine embeddings of local neighbourhoods to build a good global representation. Work in this vein has already been conducted in matrix factorisation in the context of recommender systems \citep{lee2013local, mackey2011divide}.

\bibliographystyle{apalike}
\bibliography{triangle_density}

\newpage

\onecolumn
\appendix
\section{APPENDIX}

\subsection{Technical details of construction of manifolds}

Suppose we have a symmetric positive-definite kernel $f: \*Z_n \times \*Z_n \rightarrow [0,1]$, where $\*Z_n \subset \R^d$ is a compact topological manifold, such as the unit cube or sphere. Also suppose that the latent positions of the graph are from a probability distribution $G_n$ on $\*Z_n$. Then nodes $i$ and $j$, with latent positions $z_i$ and $z_j$, connect with probability $f(z_i,z_j)$.

To characterise the sparsity and triangle density of the graph, observe that
\begin{align}
  \rho_n &= \iint f(x,y)  \rd G_n(x) \rd G_n(y),\nonumber\\
  \Delta_n &= \Theta(n^2) \iiint f(x,y) f(y,z) f(z,x) \rd G_n(x) \rd G_n(y)\rd G_n(z).\label{eq:Delta_n_f^3}
\end{align}

Now, consider the operator:
\[\*A g(x) = \int f(x,y) g(y) \rd G_n(y),\]
 on functions $g$ which are square-integrable with respect to the measure $G_n$. The operator $\*A$ is self-adjoint since $f$ is symmetric, and Hilbert-Schmidt, since $0\leq f(x,y)\leq 1$. To compute the triangle density we will need to work with the eigenvalues of $\*A$, denoted $\lambda_1, \lambda_2, \ldots$, which are nonnegative since $f$ is positive-definite. The dependence of these eigenvalues and $\*A$ on $n$ is not shown in the notation.

 The following lemma makes a crucial connection between the eigenvalues of $\*A$ and the expected triangle density $\Delta_n$ -- recall \eqref{eq:Delta_n_f^3}.
\begin{lemma}
\label{lem:sum_lambda}
  If $f$ is Lipschitz continuous in each variable,
  \[\iiint f(x,y) f(y,z) f(z,x) \rd G_n(x) \rd G_n(y)\rd G_n(z) = \sum_{k=1}^\infty \lambda_i^3.\]
\end{lemma}
Thus by understanding the behaviour of $\sum_{k=1}^\infty \lambda_i^3$  we can understand the behaviour of the expected triangle density as $n\to\infty$. This is a key part of our proofs for our explicit examples.

Now, let us consider the Mercer representation (see e.g. \citet{steinwart2008support}) of our kernel $f$:
\[f(x,y) = \sum_{k=1}^{\infty} \gamma_k u_k(x) u_k(y),\]
where $u_1, u_2, \ldots$ and $\gamma_1, \gamma_2, \ldots$ are respectively the orthonormal eigenfunctions and corresponding (positive) eigenvalues of $f$ (their dependence on $n$ implicit), with respect to Lebesgue measure on some ball $\*B_n \subset\R^d$ containing $\*Z_n$, and the convergence of the sum is absolute and uniform. Finally, we set:
$$
\phi_n(z)=[\gamma^{1/2}_1 u_1(z)\;\, \gamma^{1/2}_2 u_2(z)\;\, \cdots]^\top,
$$
that is, $\phi_n(z)$ is a vector in $\mathbb{R}^{\mathbb{N}}$, and as a consequence of this definition,  $f(x,y) = \langle \phi_n(x), \phi_n(y) \rangle$ for all $x,y \in \*Z_n$. Hence, the graph defined by this kernel also defines and infinite-dimensional RDPG, with latent positions $\phi_n(z_i)$. It is neither obvious nor generally true that a kernel produces a map $\phi_n$ which is homeomorphic and non-distortive, but our examples will use kernels that do. 

We can then build a manifold $\*M_n$ using a homeomorphism $\*M_n := \phi_n(\*Z_n)$ which is non-distortive: the Euclidean length of a curve $\eta(t)$ on $\*Z_n$ is equal to the (infinite dimensional) Euclidean length of the image of this curve $\phi_n\{\eta(t)\}$ on $\*M_n$. If $\*Z_n = [0,1]^2$, for example, we can think of a sheet of paper which is being bent or folded, but not cut or stretched.

As a result, the set $\*M_n$ is a topological manifold of the same dimension as $\*Z_n$, and from a probability distribution $G_n$ on $\*Z_n$, which we push forward to $\*M_n$ using $\phi_n$, we obtain an identical distribution $F_n$ on $\*M_n$, when viewed as a density with respect to intrinsic Euclidean volume on the manifold. In simpler terms, a uniform (or normal) distribution on $\*Z_n$ gives a uniform (respectively, normal) distribution on $\*M_n$. 

Therefore, we are in the setting of an infinite-dimensional RDPG, with latent positions on a low-dimensional manifold, and, as long as we have access to the eigenvalues of the $\mathcal{A}$, we can calculate the expected triangle density of the resulting graph.

\subsection{Proof of Lemma \ref{lem:sum_lambda}}

Note that since it is bounded, the kernel $f$ is Hilbert-Schmidt with respect to $G_n$ in the sense that:
\[\int |f(x,y)|^2 dG_n(x) dG_n(y) < \infty,\]
since the above is in fact bounded by 1, and so the eigenvalues $\lambda_1, \lambda_2 , \ldots$ are square-summable .

From \cite{brislawn1991traceable}, we have the following:
\begin{lemma}
\label{trace-identity}
Let $\mu$ be a $\sigma$-finite Borel measure on a second-countable space, X, and let $K$ be a trace class operator on $L^2(X, \mu)$. If the kernel $K(x,x)$ is continuous at $(x,x)$ for almost every $x$ then
$$ \textnormal{tr } K = \int K(x,x) d\mu (x). $$
\end{lemma}

Let us denote by  $\mathcal{A}^3$ the operator
\[\mathcal{A}^3g(u) = \int \left(\int f(u,y)f(y,z)f(z,x) dG_n(y) dG_n(z) \right) g(x)dG_n(x) ,\]
with eigenvalues $\lambda_i^3$. Since the $\lambda_i$ are square-summable, it follows that $\mathcal{A}^3$ is trace-class. 

Now, assuming that $f$ is Lipschitz continuous, we will show that the Lipschitz continuity of the kernel of $\mathcal{A}^3$, $$f^3(u,x) = \int f(u,y)f(y,z)f(z,x) dG_n(y) dG_n(z),$$ on the diagonal follows. Let $K > 0$ be such that $|f(x_1,y) - f(x_2, y)| < K|x_1 - x_2|$ for all $x_1, x_2, y \in \*Z_n$. Then, with $f^2(x,y) = \int f(x,z)f(z,y) \rd G_n(z)$, we have
\begin{align*}
|f^2(x,x) - f^2(y,y)| &= \left|\int f(x,z)f(z,x) - f(y,z)f(z,y) \rd G_n(z) \right| \\
& \leq \int \left| f(x,z)f(z,x) - f(x,z)f(z,y) + f(x,z)f(z,y) - f(y,z)f(z,y) \right| \rd G_n(z) \\
& \leq \int f(x,z)|f(z,x) - f(z,y)| + f(z,y)|f(x,z) - f(y,z)|\rd G_n(z) \\
& \leq 2K|x -y|.
\end{align*}
Hence $f^2$ is Lipschitz continuous on the diagonal. By a similar argument, we can see that $f^2$ is Lipschitz continuous in each argument:
\begin{align*}
|f^2(x,y_1) - f^2(x,y_2)| &= \left|\int f(x,z)f(z,y_1) - f(x,z)f(z,y_2) \rd G_n(z) \right| \\
&\leq \int f(x,z) |f(z, y_1) - f(z,y_2)|  \rd G_n(z) \\
&\leq K|y_1 -y_2|,
\end{align*}
and therefore it follows by induction that $f^3$ is Lipschitz continuous on the diagonal.

Furthermore, since $G_n$ is a probability measure it is $\sigma$-finite, hence we can apply Lemma \ref{trace-identity} to $\mathcal{A}^3$ to get:
\[\iiint f(x,y) f(y,z) f(z,x) dG_n(x) dG_n(y) dG_n(z) = \sum_{i=1}^{\infty} \lambda_i^3.\] \qed

\subsection{Proof of Lemma \ref{GRBF_lemma}}

Let $f(x,y) = \exp\{-\frac{1}{2}| x - y|^2\}$, be the Gaussian radial basis function (RBF) and let $\bar{G}_n$ be a Gaussian distribution with mean zero and variance $\sigma^2_n$. 

\begin{proposition}\label{prop:phi_RBF}
The map $\phi^{\text{RBF}}_n$ associated with $f$ is a non-distortive homeomorphism.
\end{proposition}

The eigenvalues of the integral operator associated with $f$ and $\bar{G}_n$ are given explicitly by \cite{shi2009data}:
$$ \gamma_k = \sqrt{\frac{2}{1 + 2\sigma_n^2 + \sqrt{1 + 4\sigma_n^2}}} \left( \frac{2\sigma_n^2}{1 + 2\sigma_n^2 + \sqrt{1 +4\sigma_n^2}} \right)^{k-1} $$
for $k=1,2,...$ . Let $r_n = 2\sigma_n^2 / (1 + 2\sigma_n^2 + \sqrt{1 +4\sigma_n^2})$ be the ratio that is raised to $(k-1)$ in the expression above. Then using the geometric sum to infinity formula, 
\begin{equation*}
\sum_{k=1}^\infty (r_n^3)^{k-1} = \frac{1}{1 - r_n^3} 
= \frac{(1 + 2\sigma_n^2 + \sqrt{1 +4\sigma_n^2})^3}{(1 + 2\sigma_n^2 + \sqrt{1 +4\sigma_n^2})^3 - 8\sigma_n^6},
\end{equation*}
we see that  $\sum_{k=1}^\infty (r_n^3)^{k-1} = \Theta(\sigma_n)$ and thus $\sum_{k=1}^\infty \gamma_k^3 = \Theta(\sigma_n^{-2})$.

Now, let $G_n$ be a Gaussian distribution with mean zero and variance $\sigma^2_n$, truncated to the interval $[-t_n, t_n]$.  We let $t_n = \sigma^2_n$ so that the mass being removed shrinks to zero. Denote the eigenvalues of the integral operator associated with $f$ and $G_n$ by $\lambda_k$. Then,
\begin{align*}
\sum_{k=1}^\infty \lambda_k^3 &= \int_{-t_n}^{t_n} f(x,y)f(y,z)f(z,x) \rd G_n(x)\rd G_n(y)\rd G_n(z) \\
&= \frac{1}{(1 - 2\Phi(-t_n/\sigma_n))^3}\int_{-t_n}^{t_n} f(x,y)f(y,z)f(z,x) \rd\bar{G}_n(x)\rd\bar{G}_n(y)\rd\bar{G}_n(z),
\end{align*}
where $\Phi$ is the standard normal cumulative distribution function.

Using that
$$ \sum_{k=1}^\infty \gamma_k^3 = \int_{-\infty}^\infty f(x,y)f(y,z)f(z,x) \rd\bar{G}_n(x)\rd\bar{G}_n(y)\rd\bar{G}_n(z) = \Theta(\sigma_n^{-2}), $$
we will show that $\sum_{k=1}^\infty \lambda_k^3 = \Theta(\sigma_n^{-2})$. Let 
\begin{align*}
    I_{t_n} &= \int_{-t_n}^{t_n}f(x,y)f(y,z)f(z,x) \rd\bar{G}_n(x)\rd\bar{G}_n(y)\rd\bar{G}_n(z), \\
    I_{\text{res}} &= \int_{-\infty}^{-t_n}f(x,y)f(y,z)f(z,x) \rd\bar{G}_n(x)\rd\bar{G}_n(y)\rd\bar{G}_n(z) + \int_{t_n}^{\infty}f(x,y)f(y,z)f(z,x) \rd\bar{G}_n(x)\rd\bar{G}_n(y)\rd\bar{G}_n(z),
\end{align*}
so that $\sum_{k=1}^\infty {\gamma_k}^3 = I_{t_n} + I_{\text{res}},$ and
$$ \sum_{k=1}^{\infty} \lambda_k^3 = \frac{1}{(1 - 2\Phi(-t_n/\sigma_n))^3} I_{t_n}. $$
First, we will show that $I_\text{res} = o(\sigma_n^{-2})$ and therefore $I_{t_n} = \Theta(\sigma_n^{-2})$. Since $f$ is bounded above by 1, we have that 
\begin{align*}
    I_\text{res} &\leq \int_{-\infty}^{-t_n} \rd\bar{G}_n(x)\rd\bar{G}_n(y)\rd\bar{G}_n(z) + \int_{t_n}^{\infty} \rd\bar{G}_n(x)\rd\bar{G}_n(y)\rd\bar{G}_n(z) \\
    &\leq 3\int_{-\infty}^{-t_n} \rd\bar{G}_n(x) + 3\int_{t_n}^{\infty} \rd\bar{G}_n(x) \\
    &= 6 \Phi \left( - \frac{t_n}{\sigma_n}\right)
\end{align*}
Plugging in $t_n = \sigma_n^2$, we get that 
$I_\text{res} \leq 6 \Phi (-\sigma_n) \leq C\exp(-v\sigma_n^2)$, for positive constants $C,v,$. Hence, $I_\text{res} = o(\sigma_n^{-2})$. Now, with $t_n = \sigma_n^2$, we have that $(1 - 2\Phi(-t_n/\sigma_n))^{-3} = (1 - 2\Phi(-\sigma_n))^{-3} \to 1$ as $n \to \infty$. Therefore, there exists an $N \in \mathbb{N}$ such that for all $n \geq N$, $(1 - 2\Phi(-\sigma_n))^{-3} \in [1,1.1].$ It follows that $ \sum_{k=1}^{\infty} \lambda_k^3 = \Theta(\sigma_n^{-2})$. Finally, from Lemma 2, we know that
$$ \Delta_n = \Theta(n^2) \sum_{k=1}^\infty \lambda_k^3, $$
and therefore, $\Delta_n = \Theta(n^2 \sigma_n^{-2})$.

Now, we will show that the sparsity factor, $\rho_n$, for the graph generated by $f$ and $G_n$ is $ \Theta(\sigma_n^{-1})$. The sparsity factor for the graph generated by $f$ and $\bar{G}_n$ is given by
\begin{align*}
\bar{\rho}_n &= \int \int f(x,y) d\bar{G}_n(x) d\bar{G}_n(y) \\
&= \frac{1}{2\pi \sigma_n^2} \int \int \exp \left( \frac{-(x-y)^2}{2} \right) \exp \left( \frac{-(x^2 +y^2)}{2\sigma_n^2} \right) dx dy \\
&= \frac{1}{2\pi \sigma_n^2} \int
\exp \left( \frac{-y^2(\sigma_n^2 + 1)}{2\sigma_n^2} \right)
\int \exp \left( \frac{-x^2(\sigma_n^2 + 1) + 2xy\sigma_n^2}{2\sigma_n^2} \right) dx dy \\
&= \frac{1}{2\pi \sigma_n^2} \int
\exp \left( \frac{-y^2(\sigma_n^2 + 1)}{2\sigma_n^2} \right)
\int \exp \left( \frac{-(\sigma_n^2 + 1) \left(x + \frac{y\sigma_n^2}{\sigma_n^2 +1}\right)^2 + \frac{y^2 \sigma_n^4}{\sigma_n^2 + 1}}{2 \sigma_n^2} \right) dx dy \\
&= \frac{1}{2\pi \sigma_n^2} \int 
\exp \left( -y^2 \left(  \frac{\sigma_n^2 + 1}{2\sigma_n^2} - \frac{\sigma_n^2}{2(\sigma_n^2 + 1)} \right) \right)
\int \exp \left( \frac{-(\sigma_n^2 + 1) \left(x + \frac{y\sigma_n^2}{\sigma_n^2 +1}\right)^2}{2 \sigma_n^2} \right) dx dy \\
&= \frac{1}{2\pi \sigma_n^2} \frac{\sqrt{2\pi} \sigma_n}{\sqrt{\sigma_n^2 + 1}}
\int \exp \left( -y^2 \left( \frac{(\sigma_n^2 + 1)^2 - \sigma_n^4}{2\sigma_n^2(\sigma_n^2 + 1)} \right) \right) dy \\
&= \frac{1}{\sqrt{2\pi}\sigma_n \sqrt{\sigma_n^2 + 1}} \cdot
 \frac{\sqrt{2\pi} \sigma_n \sqrt{\sigma_n^2 + 1}}{\sqrt{2\sigma_n^2 + 1}} \\
 &= \frac{1}{\sqrt{2\sigma_n^2 + 1}}.
\end{align*}
Therefore, $\bar{\rho}_n = \Theta(\sigma_n^{-1})$. Let
\begin{align*}
    I_{t_n}^\prime &= \int_{-t_n}^{t_n}f(x,y) \rd\bar{G}_n(x)\rd\bar{G}_n(y), \\
    I_{\text{res}}^\prime &= \int_{-\infty}^{-t_n}f(x,y) \rd\bar{G}_n(x)\rd\bar{G}_n(y) + \int_{t_n}^{\infty}f(x,y) \rd\bar{G}_n(x)\rd\bar{G}_n(y),
\end{align*}
so that $\bar{\rho}_n = I_{t_n}^\prime + I_{\text{res}}^\prime$ and
$$ \rho_n = \int_{-t_n}^{t_n}f(x,y) \rd G_n(x) \rd G_n(y) = \frac{1}{(1 - 2\Phi(-t_n/\sigma_n))^2} I_{t_n}^\prime. $$

Since $f$ is bounded above by 1, we have that 
\begin{align*}
    I_\text{res}^\prime &\leq \int_{-\infty}^{-t_n} \rd\bar{G}_n(x)\rd\bar{G}_n(y) + \int_{t_n}^{\infty} \rd\bar{G}_n(x)\rd\bar{G}_n(y) \\
    &\leq 2\int_{-\infty}^{-t_n} \rd\bar{G}_n(x) + 2\int_{t_n}^{\infty} \rd\bar{G}_n(x) \\
    &= 4 \Phi \left( - \frac{t_n}{\sigma_n}\right)
\end{align*}
Plugging in $t_n = \sigma_n^2$, we get that 
$I_\text{res}^\prime \leq 4 \Phi (-\sigma_n) \leq C\exp(-v\sigma_n^2)$, for positive constants $C,v,$. Hence, $I_\text{res}^\prime = o(\sigma_n^{-2})$ and thus $I_{t_n}^\prime = \Theta(\sigma_n^{-1})$. Now, with $t_n = \sigma_n^2$, we have that $(1 - 2\Phi(-t_n/\sigma_n))^{-2} = (1 - 2\Phi(-\sigma_n))^{-2} \to 1$ as $n \to \infty$. Therefore, there exists an $N \in \mathbb{N}$ such that for all $n \geq N$, $(1 - 2\Phi(-\sigma_n))^{-2} \in [1,1.1].$ It follows that $\rho_n = \Theta(\sigma_n^{-1})$. \qed

\subsection{Proof of Lemma \ref{sphere_lemma}}

Let $f(x,y) = \exp\{x^\top y\}/\exp(r_n)$ on the circle $\*Z_n = \{ x \in \mathbb{R}^2 : \|x\|=r_n \}$, with $\*B_n = \{ x \in \mathbb{R}^2 : \|x\|\leq r_n + \epsilon\}$, for some $\epsilon > 0$. 
\begin{proposition}\label{prop:phi_spherical}
The map $\phi^{\text{S}}_n$ associated with $f$ is a non-distortive homeomorphism.
\end{proposition}
The above is proved by a straightforward application of \citet[Proposition 2]{whiteley2021matrix}.

Denote the eigenvalues of the integral operator associated with $f$ and $G_n$ by $\lambda_k$. Then, by the symmetry of the uniform distribution on $S^1(r_n)$,
\begin{align*}
\sum_{k=1}^\infty \lambda_k^3 &= \int \int \int f(x,y)f(y,z)f(z,x) \rd G_n(x)\rd G_n(y)\rd G_n(z) \\
&= \int \int f(x^\prime,y)f(y,z)f(z,x)\rd G_n(y)\rd G_n(z)
\end{align*}
for any $x^\prime \in S^1(r_n)$. Let $x^\prime = (r_n,0)$, $\mathcal{R}_\varepsilon = \{ x \in \mathbb{R}^2 : d_\circ(x, x^\prime) \leq \varepsilon \}$ for some $\varepsilon > 0$ and $f^\prime(x,y) = \exp\{ -\frac{1}{2}d_\circ(x, y)^2 \}$, where $d_\circ(x,y)$ represents the geodesic distance between $x$ and $y$ on $S^1(r_n)$. Note that $f(x,y) \geq f^\prime(x,y)$ for all $x,y \in S^1(r_n)$, and thus
\begin{align*}
\sum_{k=1}^\infty \lambda_k^3 &\geq \int \int f^\prime(x^\prime,y)f^\prime(y,z)f^\prime(z,x)\rd G_n(y)\rd G_n(z) \\
& \geq \int_{\mathcal{R}_\varepsilon} f^\prime(x^\prime,y)f^\prime(y,z)f^\prime(z,x)\rd G_n(y)\rd G_n(z).
\end{align*}
Notice that this integral is equivalent to integrating over the flat approximation of $S^1(r_n) \cap \mathcal{R}_\varepsilon$ at $x^\prime=(r_n,0)$, which is simply the vertical line between $(r_n, \varepsilon)$ and $(r_n, -\varepsilon)$. In this case, the first coordinate is $r_n$ for all points we are integrating over, and the second coordinate is uniformly distributed on $[-\varepsilon, \varepsilon]$, therefore
$$ \sum_{k=1}^\infty \lambda_k^3 \geq \frac{1}{(2\pi r_n)^2} \int_{-\varepsilon}^\varepsilon \int_{-\varepsilon}^\varepsilon f(x^\prime_2,y_2)f(y_2,z_2)f(z_2,x_2)\rd y_2 \rd z_2,  $$
where $x_2^\prime, y_2$ and $z_2$ are the second coordinates of $x^\prime, y$ and $z$, respectively. It follows that,
\begin{align*}
\sum_{k=1}^\infty \lambda_k^3 &= \frac{1}{(2\pi r_n)^2} \int_{-\varepsilon}^\varepsilon \int_{-\varepsilon}^\varepsilon \exp \left( -\frac{1}{2}\left( y_2^2 + (y_2 -z_2)^2 + z_2^2 \right) \right) \rd y_2 \rd z_2 \\
&= \frac{1}{(2\pi r_n)^2} \int_{-\varepsilon}^\varepsilon \exp(-z_2^2) 
\int_{-\varepsilon}^\varepsilon \exp \left(-(y_2^2 - y_2z_2) \right) \rd y_2 \rd z_2 \\
&= \frac{1}{(2\pi r_n)^2} \int_{-\varepsilon}^\varepsilon \exp\left(-z_2^2 + \frac{z_2^2}{4}\right) 
\int_{-\varepsilon}^\varepsilon \exp \left(-\left(y_2 - \frac{z_2}{2}\right)^2 \right) \rd y_2 \rd z_2 \\
&= \frac{\sqrt{\pi}}{(2\pi r_n)^2} \int_{-\varepsilon}^\varepsilon \exp\left(-\frac{3}{4}z_2^2\right) 
\left[ \Phi\left( \sqrt{2}\left(\varepsilon - \frac{z_2}{2}\right) \right) - \Phi\left( \sqrt{2}\left(-\varepsilon - \frac{z_2}{2}\right) \right)  \right] \rd z_2,
\end{align*}
where $\Phi$ is the standard normal cumulative distribution function. Since $z_2 \in [-\varepsilon, \varepsilon]$, we can lower bound this by
\begin{align*}
\sum_{k=1}^\infty \lambda_k^3 &\geq \frac{\sqrt{\pi}}{(2\pi r_n)^2} \left[ \Phi\left( \frac{\varepsilon}{\sqrt{2}} \right) - \Phi\left( \frac{-\varepsilon}{\sqrt{2}} \right)  \right] \int_{-\varepsilon}^\varepsilon \exp\left(-\frac{3}{4}z_2^2\right) 
 \rd z_2 \\
 &= \frac{\sqrt{\pi}}{(2\pi r_n)^2} \mathbb{P}\left(|Z| \leq \frac{\varepsilon}{\sqrt{2}}\right) \int_{-\varepsilon}^\varepsilon \exp\left(-\frac{3}{4}z_2^2\right) \rd z_2,
\end{align*}
where $Z \sim \mathcal{N}(0,1)$. Then, by Chebyshev's inequality,
\begin{align*}
\sum_{k=1}^\infty \lambda_k^3 &\geq \frac{\sqrt{\pi}}{(2\pi r_n)^2} \left( 1 - \frac{2}{\varepsilon^2} \right) \int_{-\varepsilon}^\varepsilon \exp\left(-\frac{3}{4}z_2^2\right) \rd z_2 \\
&= \frac{\sqrt{\pi}}{(2\pi r_n)^2} \left( 1 - \frac{2}{\varepsilon^2} \right) \frac{2\sqrt{\pi}}{\sqrt{3}} \left[ \Phi \left( \frac{\varepsilon}{\sqrt{2/3}} \right) - \Phi \left( \frac{-\varepsilon}{\sqrt{2/3}} \right) \right] \\
&= \frac{1}{2\sqrt{3}\pi r_n^2} \left( 1 - \frac{2}{\varepsilon^2} \right) \mathbb{P}\left(|Z| \leq \frac{\varepsilon}{\sqrt{2/3}}\right) \\
& \geq \frac{1}{2\sqrt{3}\pi r_n^2} \left( 1 - \frac{2}{\varepsilon^2} \right) \left( 1 - \frac{2}{3\varepsilon^2} \right) \\
& \geq \frac{1}{2\sqrt{3}\pi r_n^2} \left( 1 - \frac{2}{\varepsilon^2} \right)^2.
\end{align*}
Letting $\varepsilon =2$, we get that 
$$ \sum_{k=1}^\infty \lambda_k^3 \geq \frac{1}{8\sqrt{3}\pi r_n^2} ,$$
and hence $\sum_{k=1}^\infty \lambda_k^3 = \Omega(r_n^{-2})$. Finally, using Lemma \ref{lem:sum_lambda}, we have $\Delta_n = \Omega(n^2 r_n^{-2})$.

Now we will show that the sparsity factor, $\rho_n$, associated with $f$ and $G_n$ is $O(r_n^{-1})$. We have that, by the symmetry of the uniform distribution on $S^1(r_n)$,
\begin{align*}
\rho_n = \int \int f(x,y) \rd G_n(x) \rd G_n(y) = \int f(x^\prime,y) \rd G_n(y) 
\end{align*}
for any $x^\prime \in S^1(r_n)$. Therefore,
\begin{align*}
\rho_n &= \int \exp \left( - \frac{||x^\prime - y||^2}{2} \right) \rd G_n(y) \\
&= \int \exp \left( \frac{-({x^\prime}^\top x^\prime + y^\top y -2 {x^\prime}^\top y)}{2} \right) \rd G_n(y) \\
&= \exp (-r_n^2) \int \exp (-{x^\prime}^\top y) \rd G_n(y).
\end{align*}
Now, letting $x^\prime = r_n e_1$, where $e_1 = (1,0)$, and making the change of variable $u = y/r_n$, so that $u$ is uniformly distributed on the unit circle $S^1(1)$, we get
$$\rho_n = \frac{\exp(-r_n^2)}{r_n} \int \exp (-r_n^2 e_1^\top u) \rd G(u)$$
where $G$ is the uniform distribution on $S^1(1)$. Multiplying and dividing by $I_0(r_n^2)$, where $I_0$ denotes the modified Bessel function of the first kind at order 0, we get that the expression inside the integral is the probability density of the von Mises-Fisher distribution with mean direction $-e_1$ and concentration parameter $r_n^2$:
\begin{align*}
\rho_n &= \frac{\exp(-r_n^2)I_0(r_n^2)}{r_n} \int \exp \frac{1}{I_0(r_n^2)}(-r_n^2 e_1^\top u) \rd F(u) \\
&= \frac{\exp(-r_n^2)I_0(r_n^2)}{r_n}.
\end{align*}
It has been shown that $I_0(x) < \cosh x$ \citep{luke1972}, and using that $\cosh x < \exp (x)$, we get the upper bound $\rho_n < r_n^{-1} $. It follows that $\rho_n = O(r_n^{-1})$. \qed

\subsection{Proof of Proposition \ref{prop:phi_RBF}}
In order to establish that $\phi^{\text{RBF}}_n$ is a homeomorphism, we need to show it is continuous, injective, and has a continuous inverse on its image. 

The continuity of $\phi^{\text{RBF}}_n$ follows from the continuity of $(x,y)\mapsto f(x,y)$ combined with the identities:
\begin{align*}
\|\phi^{\text{RBF}}_n(x)-\phi^{\text{RBF}}_n(y)\|^2 &=\|\phi^{\text{RBF}}_n(x)\|^2 + \|\phi^{\text{RBF}}_n(y)\|^2
- 2\langle\phi^{\text{RBF}}_n(x),\phi^{\text{RBF}}_n(y)\rangle\\
&=f(x,x)+f(y,y)-2f(x,y).
\end{align*}
The proof of injectivity is by contradiction. Assume there exists $x, y\in\*Z_n$ such that $x\neq y$ and $\phi^{\text{RBF}}_n(x)=\phi^{\text{RBF}}_n(y)$. Then $1=f(x,x)=\langle\phi^{\text{RBF}}_n(x),\phi^{\text{RBF}}_n(x)\rangle=\langle\phi^{\text{RBF}}_n(x),\phi^{\text{RBF}}_n(y)\rangle = f(x,y)= \exp\{-\frac{1}{2}|x-y|^2\}\neq 1$, 
giving a contradiction as required.

Since $\*Z_n$ is compact, a general result concerning injective maps on compact domains, \cite{sutherland2009introduction}[Prop 13.26], implies that the inverse of $\phi^{\text{RBF}}_n$ on its image must be continuous.  Thus we have established $\phi^{\text{RBF}}_n$ is a homeomorphism.

Now let $\eta:[0,1]\to\*Z_n$ be a curve, i.e., a continuously differentiable function.  The length of the image of this curve by $\phi^{\text{RBF}}_n$, i.e. the length of  $t\in[0,1]\mapsto\phi^{\text{RBF}}_n\{\eta(t)\}\in\mathcal{M}_n$ is:
\begin{equation}\label{eq:curve_length}
    \int_{0}^1\left\|\frac{\mathrm{d}}{\mathrm{d}t}\phi^{\text{RBF}}_n\{\eta(t)\}\right\|\mathrm{d}t.   
\end{equation}

By \cite{steinwart2008support}[Lem 4.34], $\phi^{\text{RBF}}_n$ is continuously differentiable, hence by the chain rule:
$$
\frac{\mathrm{d}}{\mathrm{d}t}\phi^{\text{RBF}}_n\{\eta(t)\} = \frac{\mathrm{d}\phi^{\text{RBF}}_n}{\mathrm{d}x}\{\eta(t)\} \frac{\mathrm{d}\eta}{\mathrm{d}t}
$$
and
\begin{align*}
\left\|\frac{\mathrm{d}}{\mathrm{d}t}\phi^{\text{RBF}}_n\{\eta(t)\}\right\|^2&=\left\langle\frac{\mathrm{d}}{\mathrm{d}t}\phi^{\text{RBF}}_n\{\eta(t)\},\frac{\mathrm{d}}{\mathrm{d}t}\phi^{\text{RBF}}_n\{\eta(t)\}\right\rangle\\
&=\left\langle\frac{\mathrm{d}\phi^{\text{RBF}}_n}{\mathrm{d}x}\{\eta(t)\},\frac{\mathrm{d}\phi^{\text{RBF}}_n}{\mathrm{d}x}\{\eta(t)\}\right\rangle \left|\frac{\mathrm{d}\eta}{\mathrm{d}t}\right|^2\\
&=\left.\frac{\partial^2f}{\partial x \partial y}\right|_{\eta(t),\eta(t)}\left|\frac{\mathrm{d}\eta}{\mathrm{d}t}\right|^2.
\end{align*}
For the RBF kernel in question, $\left.\frac{\partial^2 f}{\partial x \partial y}\right|_{\eta(t),\eta(t)} = 1$, and substituting into \eqref{eq:curve_length}, we find:
$$
\int_{0}^1\left\|\frac{\mathrm{d}}{\mathrm{d}t}\phi^{\text{RBF}}_n\{\eta(t)\}\right\|\mathrm{d}t   =\int_{0}^1 \left|\frac{\mathrm{d}\eta}{\mathrm{d}t}\right| \mathrm{d}t, 
$$
the right hand side of which is the Euclidean length of $\eta$. Hence $\phi^{\text{RBF}}_n$ is non-distortive, as required.

\subsection{Proof of Lemma \ref{lem:gaps}}
Let $\phi_n$ be a homeomorphism satisfying $\*M_n = \phi_n([0,1])$. Define $X_{(0)}:= \phi_n(0), X_{(n+1)} := \phi_n(1)$ and
\[X_{(k)} = X_{\ell}: \text{$\phi_{n}^{-1}(X_\ell)$ is the $\ell$-th smallest of $\phi_{n}^{-1}(X_1),\ldots, \phi_{n}^{-1}(X_n)$}, \quad k \in \{1, \ldots, n\}.\]
Then 
\begin{align*}
    H_n := d_H(\{X_i\}, \*M_n) &= \max \left\{(d_\circ(X_{(1)},X_{(0)}), d_\circ(X_{(2)}, X_{(1)})/2, \ldots, d_{\circ}(X_{(n)}, X_{(n-1)})/2, d_\circ (X_{(n+1)},X_{(n)})\right\}\\
    2 H_n &\geq \underset{0 \leq k \leq n}{\max}\left\{d_\circ(X_{(k+1)},X_{(k)})\right\} =: M_n
\end{align*}
Pick $\epsilon > 0$. Then
\begin{align*}
    \Prob(H_n \geq \epsilon) & \geq \Prob(M_n \geq 2 \epsilon)
    \geq \Prob\left\{\frac{n}{v_n} M_n - \log(n) \geq \frac{2 n \epsilon}{v_n}- \log(n)\right\}
    \rightarrow S\left\{\frac{2 n \epsilon}{v_n}- \log(n)\right\},
\end{align*}
by a classical result on uniform spacings \citep{devroye1981laws}, where $v_n = d_\circ\{\phi_n(1),\phi_n(0)\}$ and $S(x) = 1-\exp(-e^{-x})$ is the survivor function of the standard Gumbel distribution. Hence, $\Prob(H_n \geq \epsilon) \rightarrow 0$ implies $n/v_n = \Omega\{\log(n)\}$.

Now, pick any $r$ satisfying $0 < r < \sqrt{c}$ and consider any $x \in \*M_n$. Then,
\begin{align*} n \rho_n(x) &:= n \int \langle x,y \rangle \rd F_n(y),\\
& \geq n \E\left\{\langle x, X\rangle \mid X \in B_r(x)\right\} \Prob\{X\in B_r(x)\},\\
& \geq \frac{n r}{v_n} \inf\{\langle x,y \rangle: y \in B_r(x)\},\\
& \geq \frac{n r}{v_n}(c^2 -  c r), 
\end{align*}
where $X \sim F_n$, $B_r(x)$ denotes a Euclidean ball of centre $x$ and radius $r$, we have used that geodesic distance is larger than Euclidean distance in the third line, and
$\langle x, y\rangle = (\langle x, x\rangle + \langle y, y\rangle - \|x-y\|^2)/2 \geq (c^2 + (c-r)^2 - r^2)/2 = c^2 -  c r$ in the fourth. As a result, 
\[n \rho_n = n \int \rho_n(x) \rd F_n(x) \geq \frac{n r}{v_n}(c^2 -  c r) = \Omega\{\log(n)\}.\]

\subsection{The graphon model}
Let node representations $Z_i \overset{\text{i.i.d.}}{\sim} \mathcal{U}[0,1]$ and $\rho_n > 0$ be monotone non-increasing. Suppose the probability of an edge between nodes $i$ and $j$ is modelled by
\[ p_{ij} = \rho_n g(Z_i, Z_j),\]
where $g: [0,1] \times [0,1] \to [0,1]$ is a symmetric function such that $\iint g(x,y) \rd x \rd y = 1$.

The triangle density $\Delta_n$ of the resulting graph is defined by
\begin{align*}
 \Delta_n &= \frac{(n-1)(n-2)}{6} \iiint_{[0,1]^3} \rho_n g(x,y) \rho_n g(y,z) \rho_n g(z,x) \rd x \rd y \rd z \\
 &= \Theta(n^2) \rho_n^3 \iiint_{[0,1]^3} g(x,y) g(y,z) g(z,x) \rd x \rd y \rd z \\
 &\leq \Theta(n^2) \rho_n^3
\end{align*}
Therefore $\Delta_n \to 0$ if $\rho_n = o(n^{-2/3})$. Hence, in a sparse regime where average degree is constant with the respect to $n$, i.e. $\rho_n = o(1/n)$, the triangle density $\Delta_n \to 0$.

\subsection{Laplacian spectral embedding}
We have referred to spectral embedding in various places in the main document. Given an undirected graph whose adjacency matrix $\+A$ has eigendecomposition $\+A = \sum_i \lambda_i u_i u_i^\top$, its \emph{adjacency spectral embedding} into $d$ dimensions is given by the rows of
\begin{equation*}
    \hat{\+X} := (|\lambda_1|^{1/2} u_1, \ldots, |\lambda_d|^{1/2} u_d).
\end{equation*}
The point cloud $\hat{\+X}$ can be interpreted as a geometric representation of the best rank-$d$ approximation of $\+A$, in the sense that, providing $\lambda_1,\ldots,\lambda_d > 0$ (as is generally assumed throughout this paper),
\begin{equation*}
    \hat{\+X} = \argmin_{\+X : \rank(\+X) \leq d} \|\+X\+X^\top - \+A\|_\text{F}.
\end{equation*}

The matrix input may also be rectangular (e.g. the `core-periphery slice' (Section \ref{sec:implication} - local embedding)), in which case, singular values and left singular vectors replace eigenvalues and eigenvectors respectively.

The graph Laplacian matrix is defined as $\+L = \+D^{-1/2}\+A\+D^{-1/2}$ where $\+D$ is the diagonal matrix containing the node degrees (the row-sums of $\+A$). An alternative spectral embedding is the \emph{Laplacian spectral embedding} defined analogously to above, with $\+A$ replaced with $\+L$.
Figures~\ref{fig:manifolds} and \ref{fig:svs_and_triangles} employ adjacency spectral embedding while Figure~\ref{fig:cp_rocs} employs Laplacian spectral embedding.

\subsection{Proof of Theorem~\ref{thm:subgraph_density}}
Let $M_n$ denote the number of latent positions in $B_{r_n}(x)$. Then $M_n$ follows a Binomial distribution with success probability exceeding $\frac{C}{v_n} r^d$, and so $\E(M) = \Omega(n r^d/v_n)$.

The expected degree of the core is
\[\sum_{m=0}^\infty m \iint_{B_{r_n}(x)} \langle y,z \rangle \rd \bar F_n(x) \rd \bar F_n(y) \Prob(M_n = m),\]
where $\bar F_n$ is the distribution $F_n$ conditioned on landing in $B_{r_n}(x)$.

First, for $y,z \in B_{r_n}(x)$,
\begin{align*}
  \langle y,z \rangle &\geq (\langle y, y\rangle + \langle z, z\rangle - \|y-z\|^2)/2 \\
  &\geq \{\langle y, y\rangle + \langle z, z\rangle - (2 r_n)^2\}/2, \\
  &\geq \langle x, x \rangle - 6 r_n^2,
\end{align*}
using the triangle inequality $\|x\| \leq \|y\| + \|x-y\|$. As a result, $ \langle y,z \rangle$ has a lower bound $\langle x, x \rangle - 6 R^2$, which is positive by the theorem assumptions, and the expected degree is larger than
 \[(\langle x, x \rangle - 6 R^2) \E(M_n) = \Omega(n r^d/v_n).\]
The triangle density of the core is 
\begin{align*}\sum_{m=0}^\infty {m \choose 3}/m \iiint_{B_{r_n}(x)} & \langle y,z \rangle \langle z,u \rangle \langle u,y \rangle\rd \bar F_n(y) \rd \bar F_n(z) \rd \bar F_n(u)\Prob(M_n = m),\\
                                                                    & \geq \left(\langle x, x \rangle - 6 R^2\right)^3 \E\left({M_n \choose 3}/M_n\right)
\end{align*}
The assumption $r_n = \omega\{(v_n/n)^{1/d}\}$ ensures that $M_n \rightarrow \infty$ in probability, so that
\[\left(\langle x, x \rangle - 6 R^2\right)^3 \E\left({M_n \choose 3}/M_n\right) = \Omega(\E\{M_n^2\}) = \Omega(n r^d/v_n)^2.\]

\subsection{Recovery of the latent positions from the core and core-periphery embeddings}

Let $\+P = \+X\+X^\top$ where $\+X \in \R^{n \times D}$ is of rank $D$ and let $\text{C}$ denote the indices of $m$ `core' nodes. Let $\+X^\text{C} \in \R^{m \times D}$ denote the rows of $\+X$ in $\text{C}$ and assume the rows of $\+X^\text{C}$ are contained within a $d$-dimensional subspace of $\R^D$, i.e. $\rank(\+X^\text{C}) = d$, where $m > d$. Let $\+P^\text{C-P slice} = \+X^\text{C} \+X^\top$ denote the $m \times n$ matrix containing the rows of $\+P$ in $\text{C}$ and let $\+P^\text{C-P slice} = \+Y^{\text{C-P}}(\+Z^{\text{C-P}})^\top$ where $\+Y^{\text{C-P}} \in \R^{m \times d}, \+Z^{\text{C-P}} \in \R^{n \times d}$, be a rank-$d$ factorisation of $\+P^\text{C-P slice}$.
Then
\begin{equation*}
  \+X^\text{C} = \+Y^{\text{C-P}} \left[ (\+Z^{\text{C-P}})^\top \+X (\+X^\top \+X)^{-1} \right],
\end{equation*}
so $\+X^{\text{C}}$ is a linear transformation of $\+Y^{\text{C-P}}$.

Let $\+P^\text{C} = \+X^\text{C} (\+X^\text{C})^\top$ denote the $m \times m$ submatrix containing the rows and columns of $\+P$ in $\text{C}$ and let $\+P^\text{C} = \+Y^\text{C} (\+Y^\text{C})^\top$, where $\+Y^\text{C} \in \R^{m \times d}$ be a rank-$d$ factorisation of $\+P^\text{C}$. There is no such relationship between $\+X^\text{C}$ and $\+Y^\text{C}$ since by assumption, $\rank(\+X^\text{C}) = d$ and so $(\+X^\text{C})^\top \+X^\text{C}$ is not invertible. To provide a concrete counter example, consider the $\+P$-matrix generated from a stochastic block model with $\+B$-matrix

\begin{equation*}
\mathbf{B} =  \begin{pmatrix}
  a & a & b \\
  a & a & c \\
  b & c & a \\
  \end{pmatrix}.
\end{equation*}

Let $\text{C}$ be the nodes in the first and second community. Then $\+X^\text{C}$ has rank 2 and $\+Y^\text{C}$ has rank 1, so they cannot be related by a linear transformation.

\subsection{Comparison of the core versus core-periphery embeddings}

\begin{figure}[H]
\centering
\includegraphics[width=\linewidth]{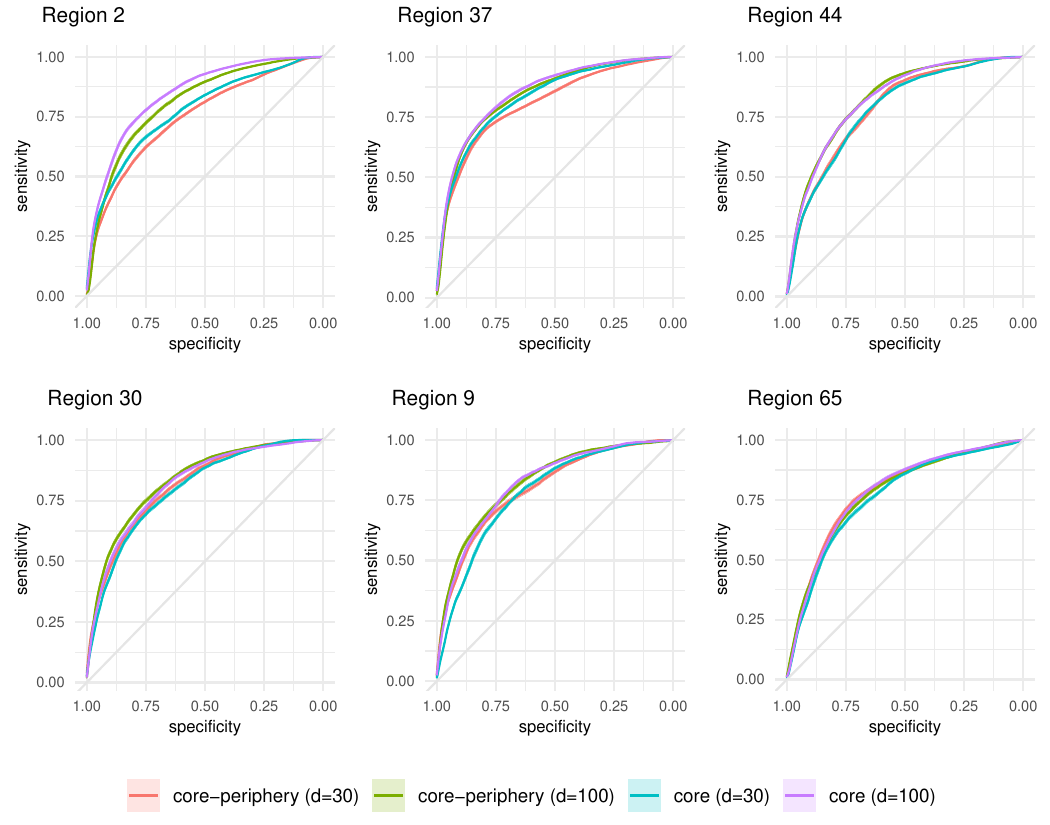}
\vspace{-7pt}
\caption{Classification of tissue types from a diffusion MRI scan. For each region, we train a classifier to predict the tissue type of each node using the core-periphery and core embeddings into dimensions $d=30$ and $d=100$. For 100 Monte-Carlo iterations, a random sample of 20\% of the positions are held out for validation and the corresponding mean receiver-operating curves, and $\pm 2$ standard errors are shown. The core-periphery embedding performs comparably to the core embedding.}
\label{fig:core_vs_cp}
\end{figure}

Figure~\ref{fig:core_vs_cp} shows the classification performance of the core-periphery and core embeddings in the real data experiment detailed in Section~\ref{sec:implication}.

\subsection{Simulations of concrete examples}

\begin{figure}[!htb]
    \centering
    \includegraphics[scale=0.4]{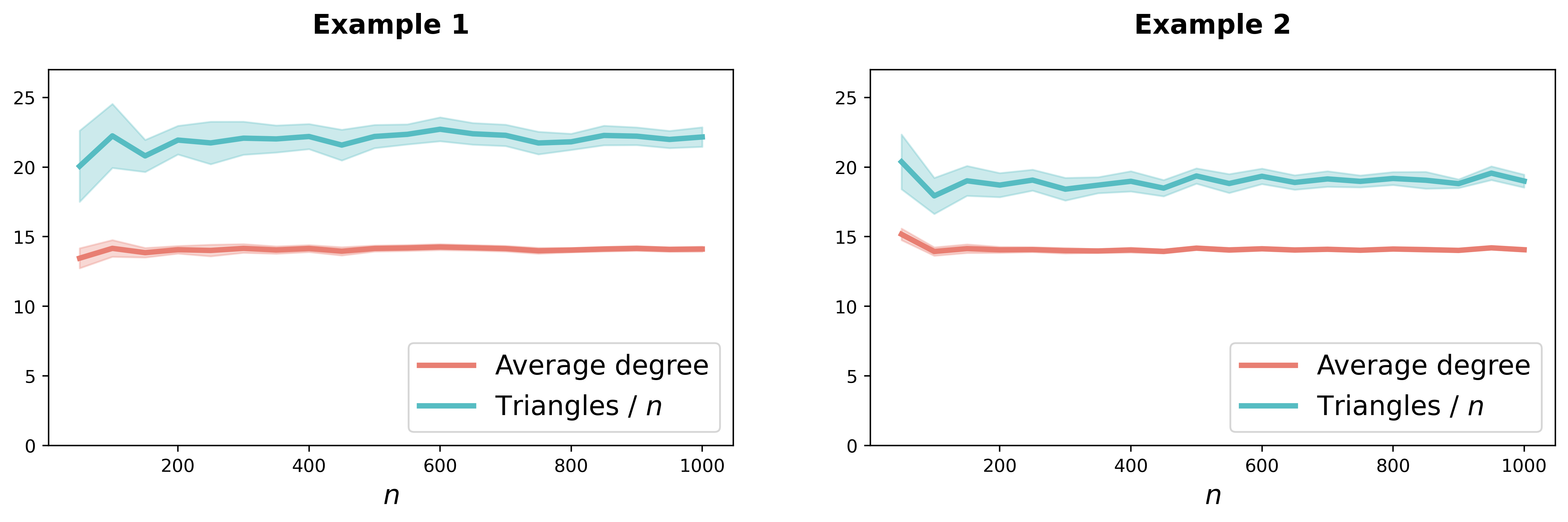}
    \caption{For each $n$, 20 graphs have been simulated from examples 1 and 2, with $\sigma_n = n/20$ and $r_n = n/50$ respectively, and the average degree ($n \rho_n$) and average triangle density ($\Delta_n$) calculated. The confidence bands shown are $\pm$ three standard errors from the mean.}
    \label{fig:constant}
\end{figure}

\citet{seshadhri2020impossibility} note that for various real-world graphs, when looking only at the subgraphs containing nodes of maximum degree $c \in [10,50]$, they get a value of $\Delta > 1$ (in many cases $\Delta > 10$). Hence, we choose values of $\sigma_n$ and $r_n$ (that are $\Theta(n)$) in our examples, which produce graphs of average degree in the same range. Figure \ref{fig:constant} shows that setting $\sigma_n = n/20$ and $r_n = n/50$ gives average degrees in this range and average $\Delta_n > 10$ for varying values of $n$. The figure also illustrates that $n \rho_n$ and $\Delta_n$ stay (close to) constant as $n$ varies.

\begin{figure}[!htb]
    \centering
    \includegraphics[scale=0.4]{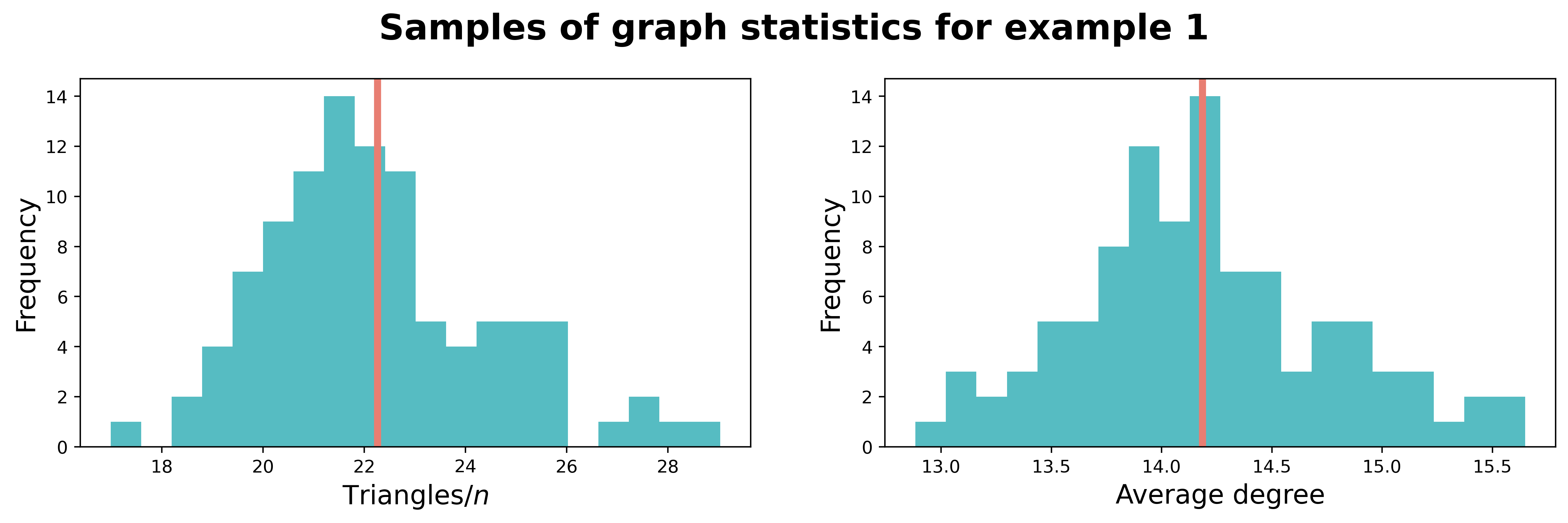}
    \caption{100 graphs of size $n=500$ have been generated from the model described in Example 1, with $\sigma_n = n/20$, and the average degree ($n \rho_n$) and triangle density ($\Delta_n$) calculated. The vertical red line represents the mean of these samples.}
    \label{fig:example1_hists}
\end{figure}

\begin{figure}[!htb]
    \centering
    \includegraphics[scale=0.4]{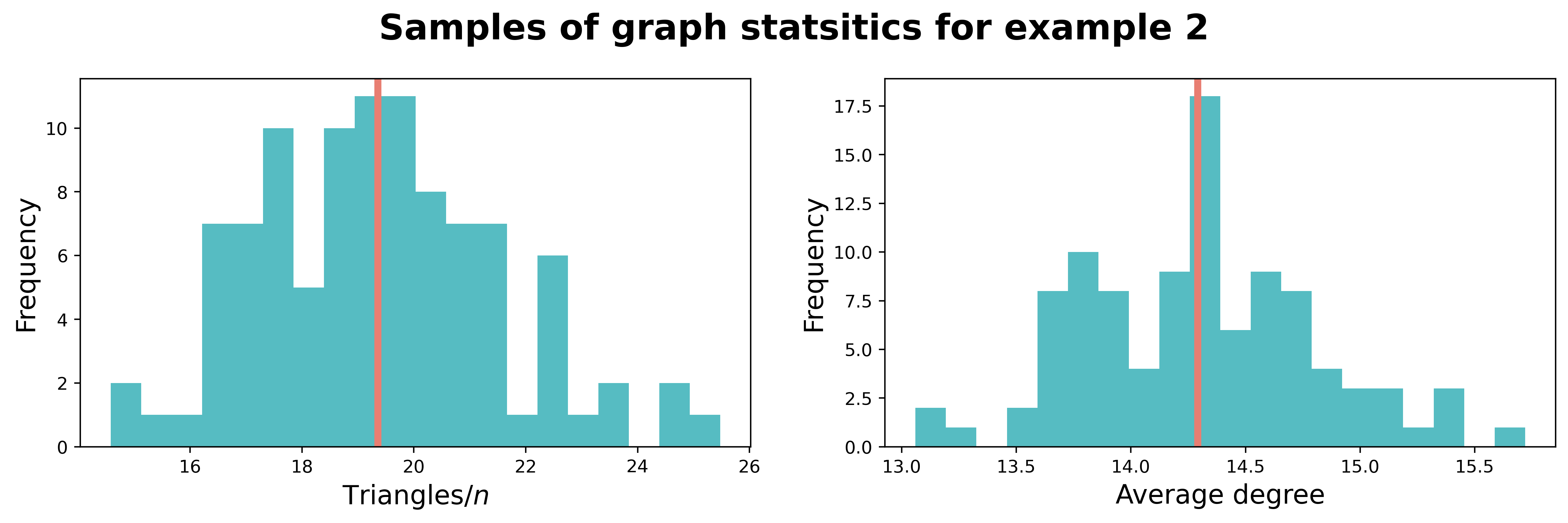}
    \caption{100 graphs of size $n=500$ have been generated from the model described in example 2, with $r_n = n/50$, and the average degree ($n \rho_n$) and triangle density ($\Delta_n$) calculated. The vertical red line represents the mean of these samples.}
    \label{fig:example2_hists}
\end{figure}

To show that the values of $\Delta_n$ and $n \rho_n$ do not vary too greatly over different simulations of graphs from the same model, Figures \ref{fig:example1_hists} and \ref{fig:example2_hists} show 100 samples from each model, again with $\sigma_n = n/20$ and $r_n = n/50$. We see that the average values (red line) correspond well to the values seen for varying $n$ in Figure \ref{fig:constant}.

\subsection{Simulations of further examples}
\label{further_examples}

To illustrate that our results apply to models other than the two that we prove theoretically, in this section we present similar results from simulations of further models.

\para{Example (3)} Let $f(x,y) = \exp\{-\frac{1}{2}| x - y|^2\}$, be the radial basis function (RBF) and $G_n$ be the uniform distribution on $[0, a_n]$.

\begin{figure}[t]
    \centering
    \includegraphics[scale=0.4]{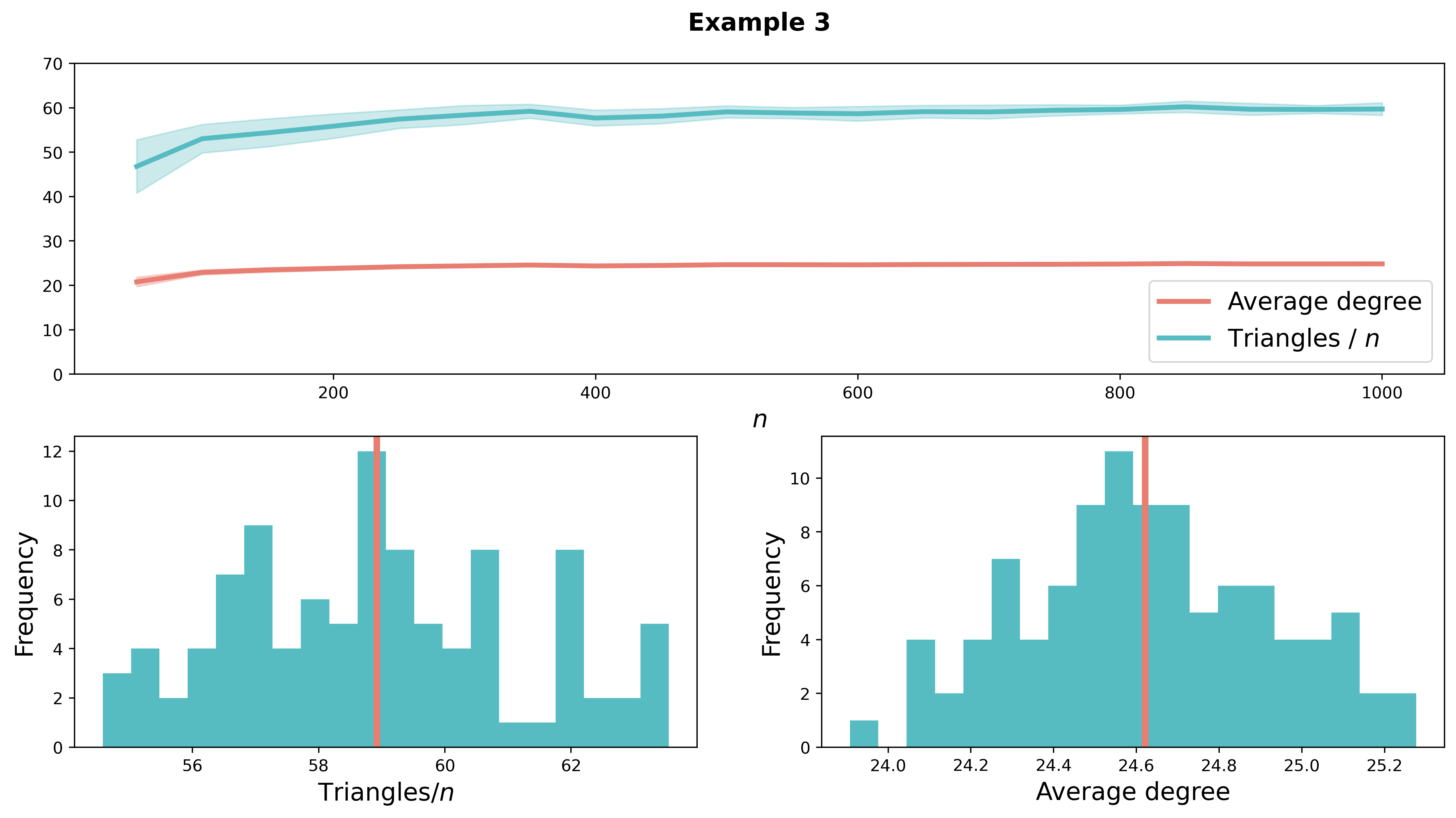}
    \caption{ In the upper plot, for each $n$, 20 graphs have been simulated from the model described in example 3, with $a_n = n/10$, and the average degree ($n \rho_n$) and triangle density ($\Delta_n$) calculated. The confidence bounds shown are $\pm$ two standard deviations from the mean. In the lower plots, the same calculations have been done on 100 simulated graphs of size $n=500$.  The vertical red line represents the mean of these samples.}
    \label{fig:example3}
\end{figure}

We show through simulations that if we let $a_n = \Theta(n)$, then we get constant average degree and constant triangle density.  This is illustrated in Figure \ref{fig:example3} by simulating graphs from the model for increasing $n$ with $a_n = n/10$.

\para{Example (4)} Let $f(x,y) = \exp\{-\frac{1}{2}|| x - y||_2^2\}$, be the radial basis function (RBF) and $G_n$ be the uniform distribution on $S^2(r_n)$.

\begin{figure}[!htb]
    \centering
    \includegraphics[scale=0.4]{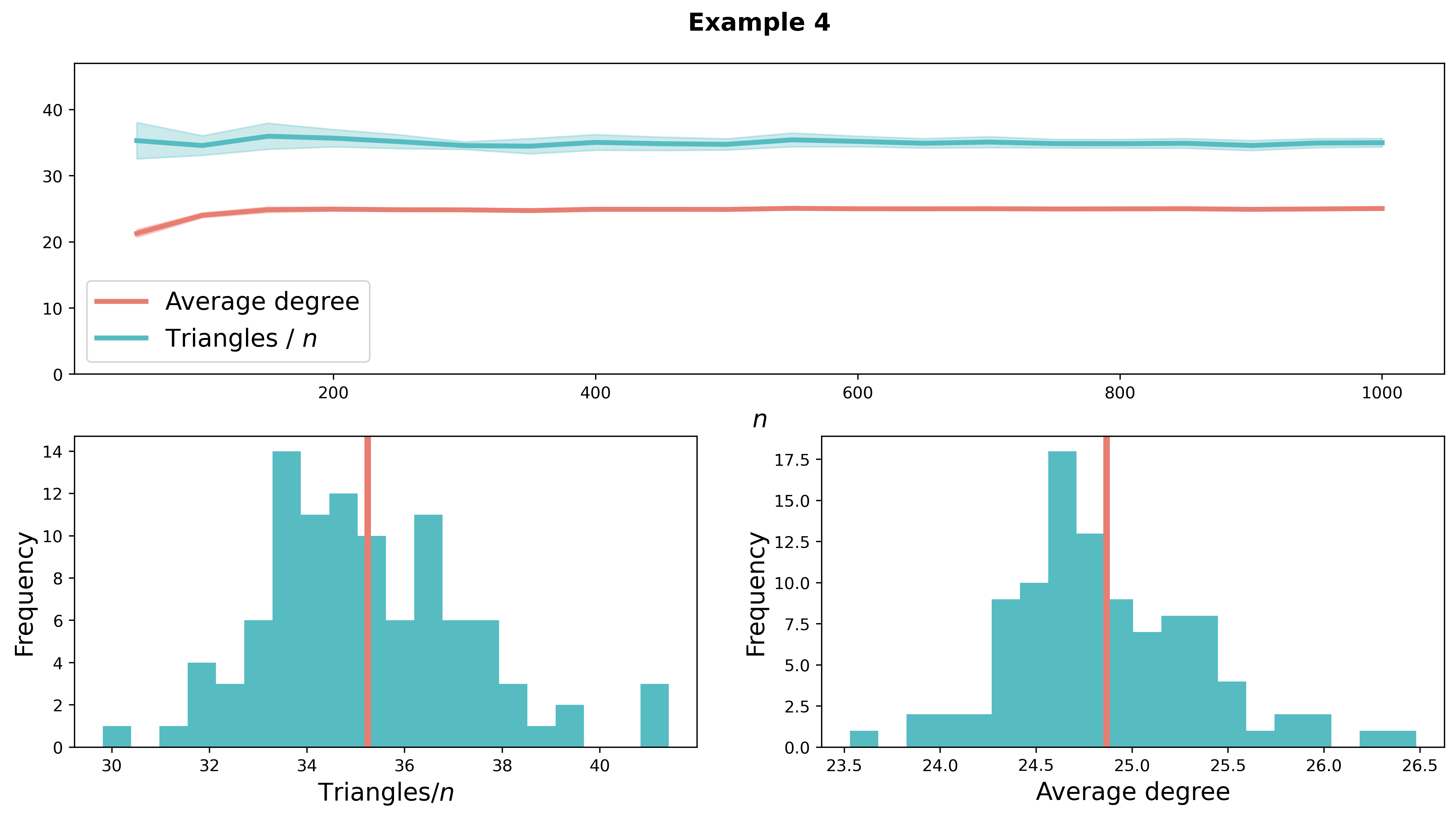}
    \caption{ In the upper plot, for each $n$, 20 graphs have been simulated from the model described in example 3, with $r_n = \sqrt{n}/10$, and the average degree ($n \rho_n$) and triangle density ($\Delta_n$) calculated. The confidence bounds shown are $\pm$ two standard deviations from the mean.  In the lower plots, the same calculations have been done on 100 simulated graphs of size $n=500$.  The vertical red line represents the mean of these samples.}
    \label{fig:example4}
\end{figure}

We show through simulations that if we let $r_n = \Theta(\sqrt{n}))$, then we get constant average degree and constant triangle density.  This is illustrated in Figure \ref{fig:example4} by simulating graphs from the model for increasing $n$ with with $r_n = \sqrt{n}/10$.

\subsection{Logistic model}

\begin{figure}[t]
    \centering
    \includegraphics[scale=0.4]{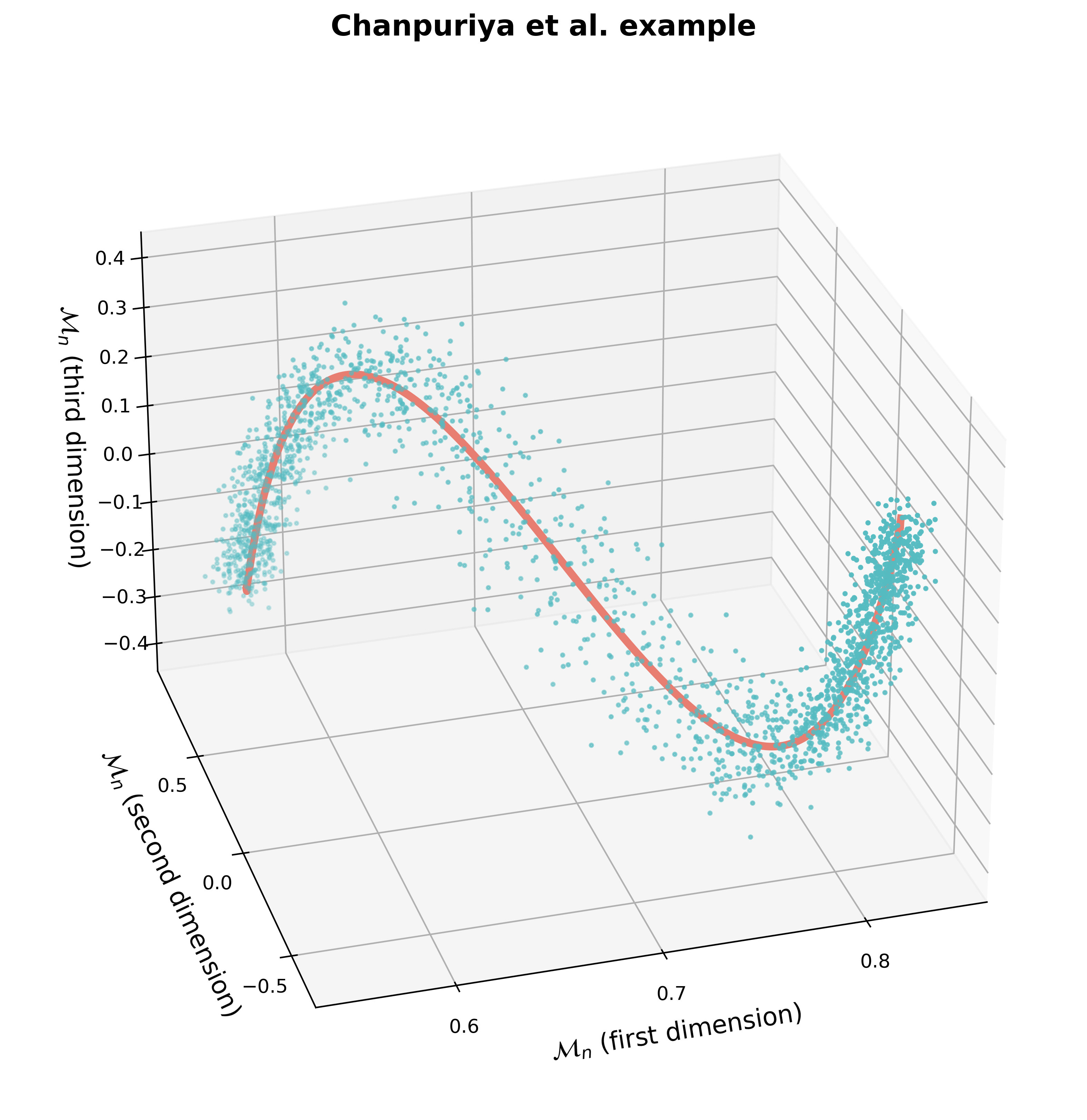}
    \caption{ The first 3 dimensions of the manifold (red curve) implied by the logistic kernel with a uniform latent position distribution $G_n$ on the interval $\mathcal{Z}_n=[-3,3]$ are shown, along with the spectral embeddings (blue points) of a graph simulated from the model and aligned using Procrustes analysis.}
    \label{fig:chanpuriya}
\end{figure}

\citet{chanpuriya2020node} propose a latent position model that they find is able to reconstruct sparse, triangle-dense graphs, where link probabilities are given by the logistic function,
$$ f(x,y) = \frac{\exp(x^\top y)}{1 + \exp(x^\top y)}. $$
As mentioned in the introduction, the effect of allowing two latent positions per node is complex, and not crucial for reproducing high triangle density. We therefore opt to present the manifold associated with the kernel above, rather than the asymmetric version in \citet{chanpuriya2020node}.
Figure~\ref{fig:chanpuriya} shows the first three dimensions of the manifold implicit in this kernel with a uniform latent position distribution (red curve) and the (Procrustes aligned) spectral embedding of a graph generated from it (blue points). The kernel is distortive (as one might expect, given it was never chosen not to be) and so the intrinsic (on-manifold) distribution $F_n$ is not uniform.

\subsection{Neighbourhood approximations}

Here we will demonstrate that the `number of shared neighbours' metric forms a reasonable surrogate for the theoretical neighbourhood described in Section \ref{sec:implication} (local embedding). The approximate neighbourhood around an arbitrary query node, $i$, is found by taking the nodes with most neighbours in common with $i$.

\begin{figure}[t]
    \centering
    \includegraphics[scale=0.8]{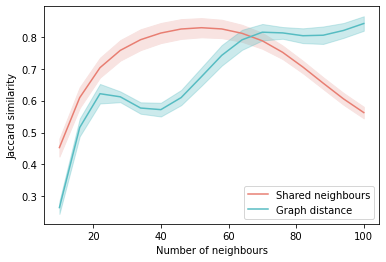}
    \caption{$y$-axis: Jaccard similarity between theoretical and approximate neighbourhoods found using two metrics: number of shared neighbours and graph distance. $x$-axis: Size of the neighbourhoods. The mean Jaccard similarity of 20 graphs simulated from the model in example (3), with $n=500$ and $a_n=n/10$, is shown for each metric, with $\pm 2$ standard deviation error bars. For each of the simulated graphs, we take the mean similarity over the neighbourhoods found for each node in the graph. The shared neighbours metric gives a mean Jaccard similarity greater than 0.7 for neighbourhoods sized between 4.4\% and 16.4\% of the size of the graph, and outperforms the graph distance metric for neighbourhoods of size less than 12.8\% the size of the graph.}
    \label{fig:shared_nhbrs}
\end{figure}

To investigate this hypothesis, we simulated a number of graphs from the model in Example (3) (Appendix \ref{further_examples}), with $n=500$ and $a_n = n/10$ so that the levels of sparsity and triangle density are in the range of real-world networks. As well as the shared neighbours metric, we also find an approximate neighbourhood based on the graph distance between nodes, i.e. the number of edges in the shortest path between nodes. Figure \ref{fig:shared_nhbrs} shows the Jaccard similarity between the approximate neighbourhoods and the theoretical neighbourhood (nodes with latent positions closest to the latent position of the query node), where the Jaccard similarity between two sets $A$ and $B$ is defined as:
$$ J(A,B) = \frac{|A \cap B|}{|A \cup B|}. $$
We see that the shared neighbours metric gives reasonable approximations to the theoretical neighbourhood for neighbourhoods of size between 5-15\% of the size of the full graph.

\end{document}